\documentclass[sigplan,screen]{acmart}
\setcopyright{none}
\renewcommand\footnotetextcopyrightpermission[1]{}

\usepackage{balance}
\usepackage{microtype}
\usepackage{booktabs}
\usepackage{subcaption}
\usepackage{graphicx}
\usepackage{xspace}
\usepackage{listings}
\usepackage{color,xcolor, colortbl}
\usepackage{amsmath}
\usepackage{pifont}

\newcommand{\fig}[1]{Figure~\ref{#1}}
\newcommand{\tab}[1]{Table~\ref{#1}}
\newcommand{\sect}[1]{Section~\ref{#1}}
\newcommand{\toolname}{\textsc{ShapeFlow}\xspace}
\newcommand{\toolnametable}{\textsc{ShapeFlow}}       
\newcommand{\var}[1]{\texttt{#1}}
\graphicspath{{figures/}}
\usepackage{graphics}
\usepackage{graphicx}
\usepackage{multirow}
\usepackage{url}
\definecolor{zhcolor}{rgb}{0.5,0,0.5}

\newcommand{\pg}[1]{\subsubsection*{\bf{#1}}}
\newcommand{\xmark}{\ding{55}}
 
\definecolor{codegreen}{rgb}{0,0.6,0}
\definecolor{codegray}{rgb}{0.5,0.5,0.5}
\definecolor{codepurple}{rgb}{0.58,0,0.82}
\definecolor{backcolour}{rgb}{1,0.96,1}
\definecolor{purple}{rgb}{0.5, 1, 0.5}

\lstdefinestyle{mystyle}{
    backgroundcolor=\color{backcolour},   
    commentstyle=\color{codegreen},
    keywordstyle=\color{magenta},
    numberstyle=\tiny\color{codegray},
    stringstyle=\color{codepurple},
    basicstyle=\ttfamily\footnotesize,
    breakatwhitespace=false,         
    breaklines=true,                 
    captionpos=b,                    
    keepspaces=true,                 
    numbers=left,                    
    numbersep=3pt,                  
    showspaces=false,                
    showstringspaces=false,
    showtabs=false,                  
    tabsize=2
}
 
\definecolor{Gray}{gray}{0.85}
\definecolor{LightCyan}{rgb}{0.88,1,1}
\definecolor{LightCyan1}{rgb}{0.6,1,1}
\newcolumntype{a}{>{\columncolor{Gray}}c}
\newcolumntype{b}{>{\columncolor{white}}c}
\newcolumntype{z}{>{\textbf}c}
\def\BibTeX{{\rm B\kern-.05em{\sc i\kern-.025em b}\kern-.08em
    T\kern-.1667em\lower.7ex\hbox{E}\kern-.125emX}}

\acmConference[]{}{}{}
\startPage{1}

\title{\toolname: Dynamic Shape Interpreter for TensorFlow}         

\author{Sahil Verma}
\affiliation{
	\institution{Department of Computer Science and Engineering\\ University of Washington\\ Seattle, USA}
}
\email{vsahil@cs.washington.edu}

\author{Zhendong Su}
\affiliation{
	\institution{Department of Computer Science\\ ETH Zurich \\Zurich, Switzerland}
}
\email{zhendong.su@inf.ethz.ch}

\begin{document}

\begin{abstract}
  We present \toolname, a dynamic abstract interpreter for TensorFlow which quickly catches tensor shape incompatibility errors, one of the most common bugs in deep learning code. 
  \toolname shares the same APIs as TensorFlow but only captures and emits tensor shapes, its abstract domain. 
  \toolname constructs a custom shape computational graph, similar to the computational graph used by TensorFlow.
  \toolname requires no code annotation or code modification by the programmer, and therefore is convenient to use.
  We evaluate \toolname on 52 programs collected by prior empirical studies to show how fast and accurately it can catch shape incompatibility errors compared to TensorFlow. 
  We use two baselines: a worst-case training dataset size and a more realistic dataset size. 
  \toolname detects shape incompatibility errors \emph{highly accurately} --- with no false positives and a single false negative --- and \emph{highly efficiently} --- with an average speed-up of 499X and 24X for the first and second baseline, respectively. We believe \toolname is a practical tool that benefits machine learning developers. 
  We will open-source \toolname on GitHub to make it publicly available to both the developer and research communities.

  
\end{abstract}

\maketitle






\section{Introduction}
\label{sec:intro}

\begin{figure}[t]
 \includegraphics[width=.9\columnwidth]{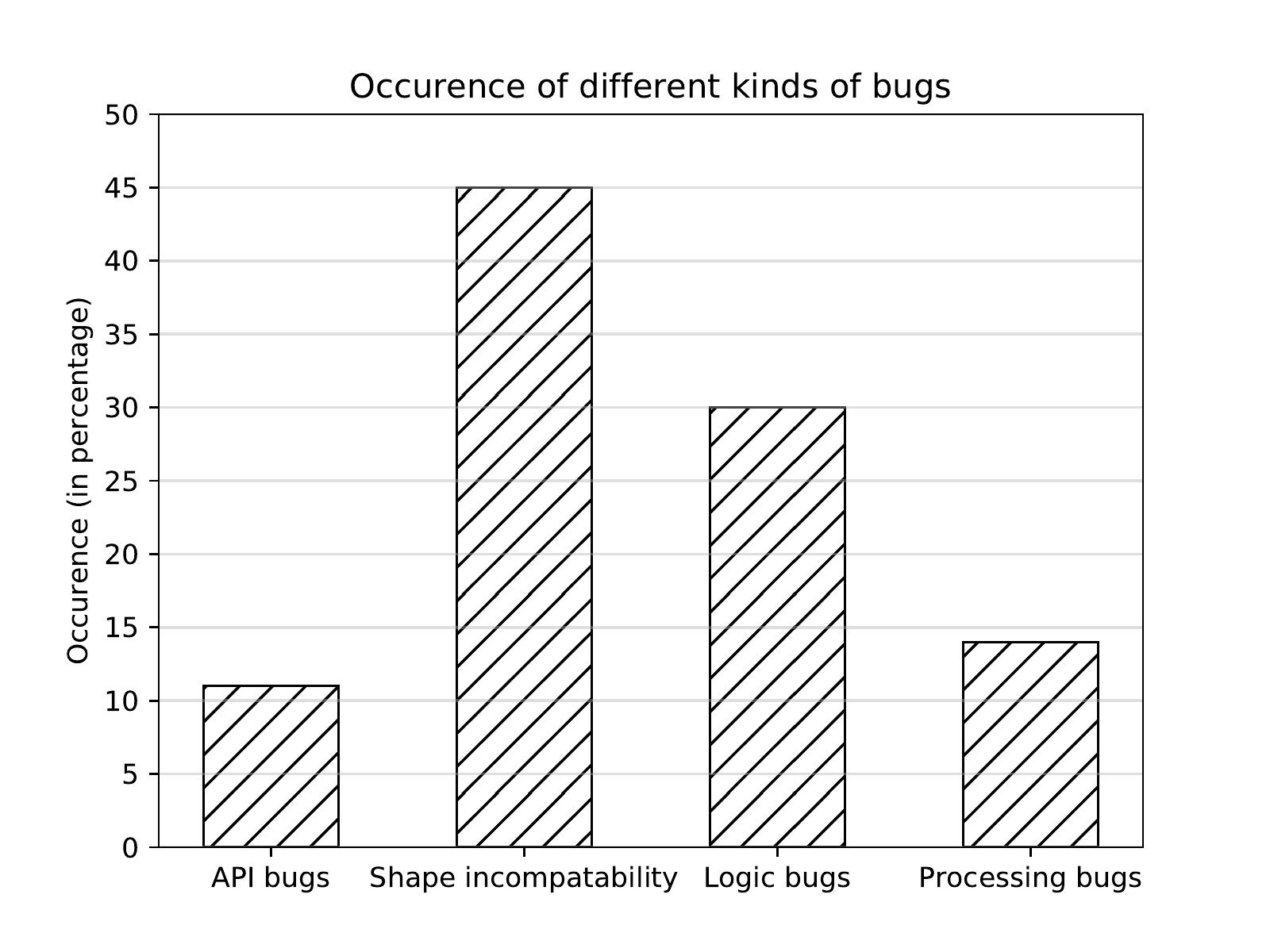}
 \caption{The prevalence of common type of bugs in TensorFlow code according to Islam et al.'s empirical study~\cite{fse-bug-collection}.}
 \label{fig:prevalence}
\end{figure}

Deep learning technologies have taken off in the last decade, and so has the code to build them.
Tools like TensorFlow~\cite{tensorFlow-paper}, Pytorch~\cite{pyTorch-paper}, Caffe~\cite{caffe-paper} and, Keras~\cite{keras}, which have been developed in the last decade, have more than 250K stars on their open-source repositories. 
Programmers use these frameworks to code their deep learning stack, and for efficient training and testing. 
Given the steep rise in deep learning use and deployment, bugs in such programs are also increasing in importance and abundance. 
\fig{fig:prevalence} shows the distribution of type of bugs that frequently occur in deep-learning code --- the information was obtained from StackOverFlow programs collected in Islam et al.'s empirical study~\cite{fse-bug-collection}.

Most of the operations in deep learning code are manipulations involving one or more tensors, which are matrices of any dimension (matrices are 2-dimensional tensors).
Each operation imposes a precondition on the shapes of the tensors that it takes as input. 
For instance, an operation might require that the larger first dimension of the two input tensors be divisible by the smaller first dimension of the second tensor.
If this precondition is met, the operation outputs a manipulated tensor whose dimension is a function of the input tensors.
When such a precondition is not met, an error is encountered. 
The error may or may not lead to a program crash.
Given the intricate API hooks provided by the libraries and the size of the code, the possibility of shape incompatibility errors is high in all stages of a deep learning pipeline, from data loading to testing.
The root causes of shape incompatibility errors include unexpected shapes of input data, typically causing a mismatch with the first layer of the model architecture or between the shapes of different architectural layers.
Unexpected input data shape errors can occur during training or testing of the model.
Even for programs that crash, encountering shape incompatibility errors can take a large amount of computation time, especially when the bug is present in later stages of the pipeline.
Lengthy pre-processing due to enormous (training and testing) dataset sizes further increases the detection time.
This can lead to a significant loss of computing and programmers' time, as all the computations performed until the program crash is wasted. 
The high prevalence of shape incompatibility errors and the typical delay in their detection motivated this work to develop a practical technique for quickly and accurately detecting such errors. An ideal technique for detecting shape incompatibility errors would:
\begin{enumerate}
 \item be accurate and fast in catching shape incompatibility errors;
 \item be capable of supporting the highly data-dependent deep learning code; and 
 \item require no code annotation.
\end{enumerate}

To the best of our knowledge, \toolname is the first such technique that helps expedite the detection of shape incompatibility errors for deep-learning code. 
The only previous work which partially attempts to address this problem is 
Ariadne~\cite{Ariadne-ibm}, a static analysis tool for machine learning code written in TensorFlow.
The programmer must annotate each line of the code in which any tensor operation takes place to indicate the expected output shape after the execution of that operation.
Annotating code is a daunting task for programmers and is a major drawback of the Ariadne approach --- its motivating example would require the programmer to annotate 16 lines of the total 36 lines in the example program.
Recently Ariadne was extended by Pythia~\cite{Pythia-ibm} which is concurrent to our work, and is a static analysis approach and therefore suffers from the standard limitations of static analysis, like it may require certain annotations; whereas \toolname takes a complementary dynamic approach.  

To this end, we propose a dynamic shape-abstracted version of TensorFlow, \toolname.
We chose TensorFlow as our deep-learning skeleton framework because it is the most popular deep learning framework~\cite{tf-popular1, tf-popular2, tf-popular3}. 
As of August 2020, 90,236 GitHub projects were using TensorFlow, which is close to twice of those (49,145 projects) using Pytorch, the second most popular deep-learning framework. 

\toolname is powered by shape-capturing APIs. 
\toolname's implementation has the same APIs as TensorFlow, but have been modified to only capture and manipulate tensor shape information.
This design has two clear advantages: 
\begin{itemize}
 \item It accelerates the code analysis for any shape incompatibility error occurring later in the deep-learning pipeline; and 
 \item Since \toolname uses Python and TensorFlow for interpretation, it does not require any annotation by its user.
\end{itemize}

We design \toolname to be dynamic because most of the deep-learning code is data-dependent, and therefore, code execution starts from inferring the input data shape and its application of operations on the data in the following layers.
If it detects any shape incompatibility errors, the execution crashes with a message similar to the one returned by TensorFlow; otherwise, it returns a ``\textit{no error detected}'' message.

We evaluate \toolname on a set of 52 programs from recent empirical studies on deep-learning bugs~\cite{issta-bug-collection, fse-bug-collection}. 
\toolname successfully detects bugs in all but one (i.e., a single false negative) of the buggy programs which have a shape incompatibility error. 
\toolname produces no false positives on the corrected versions of all the programs. 
We report the speedup \toolname achieved over TensorFlow with respect to two baselines. 
In the first baseline, we consider the case when a programmer runs their code on the entire training dataset. 
Datasets in machine learning, deep learning specifically can be enormously large, and this can significantly delay the catch of errors.
The second baseline is more realistic where the programmer runs the code on a part of the training dataset, which helps encounter any errors in the program faster, before deploying it on the entire dataset. 
This is also called sanity checking in the machine learning community.
\toolname achieved an average speedup of 499X (upto 23,532X on some benchmarks) for the first baseline and an average speedup of 24X (upto 180X on some benchmarks) for the second baseline.

To summarize, our main contributions are:
\begin{enumerate}
 \item We propose \toolname, a dynamic shape abstract version of TensorFlow for accurate and fast detection of shape incompatibility errors in deep learning code written in TensorFlow. 
 \item We implement and open-source \toolname for benefiting the community and for facilitating the review and reproduction of our results. The code is available at \url{https://github.com/vsahil/Static-Analyser_ML}. 
 \item We evaluate the effectiveness of \toolname with an extensive benchmark suite. 
\end{enumerate}

The remainder of the paper is structured as follows. 
\sect{sec:motivate} provides a motivating example to illustrate our approach. 
Computational graph for shape inference is discussed in \sect{sec:core}, and implementation details are given in \sect{sec:implement}, followed by our evaluation in \sect{sec:evaluate}. 
Finally, we discuss related work (\sect{sec:related}) and conclude (\sect{sec:conclusion}).

\section{Motivating Example}
\label{sec:motivate}

Here we illustrate shape incompatibility errors in deep-learning code using an example. 
\fig{fig:motivating_code} shows an example of a neural network architecture to be trained (and tested) on identifying hand-written digits from the MNIST dataset~\cite{lecun-mnisthandwrittendigit-2010}. 
The code has been adapted from the convolutional networks example section in the open-source code of TensorFlow~\cite{mnist-tutorial}, and modified to show only the parts of interest. 
Placeholders for images and their labels are declared on lines 18-19, which will be populated using the MNIST dataset when training or testing. 
Function \var{conv\_net} creates the convolutional neural network architecture. 
The code for training and testing module is shortened for space considerations.

To demonstrate the difference between \toolname and Ariadne, we have added the annotations in \fig{fig:motivating_code} (comments following `\#\#'), which are required by Ariadne to run and type-check tensor shapes. 
\toolname requires no such code annotations or modification.
Since all APIs in \toolname only hold shape information, i.e., consume shape of a tensor as input and emit the shape of the output, the operations are very lightweight when compared to their TensorFlow counterparts. 
In \fig{fig:motivating_code}, there is a shape incompatibility bug on line 27. 
This bug triggers after training of the model and results in a program crash. 
The programmer has interchanged the images and their labels (\colorbox{Gray}{\var{batch[0]}} and \colorbox{Gray}{\var{batch[1]}} respectively) which have different shapes, causing a program crash.
For 100 training epochs, \toolname takes 4.2s to reach the error, while TensorFlow takes 43s to reach it (a 10X speed-up). 
Ariadne was only evaluated on verifying correct code written in TensorFlow; thus, its exact behavior on buggy code is unclear. 
For Ariadne, the programmer would also need to recompute the expected shapes and modify the comments if the shape of the input data has changed because Ariadne is a static analysis tool.

\begin{figure*}[t]
 \begin{minipage}{\textwidth}
 \centering
\begin{lstlisting}[language=Python]
# Import MNIST data
from tensorflow.examples.tutorials.mnist.input_data import read_data_sets
mnist = read_data_sets("data", one_hot=False)   ## train:{images: [batch, y(28) * x(28) of channel]}
import tensorflow as tf                         ## test:{images: [batch, y(28) * x(28) of channel]}}
# Create the network architecture
def conv_net(x_dict, n_classes, dropout):       ## (x_dict: {images: [batch, y(28)*x(28) of channel]}) -> [batch, n_classes of channel]                             
    x = tf.reshape(x_dict, shape=[-1, 28, 28, 1])                 ## [batch, y(28), x(28), 1 of channel]
    conv1 = tf.layers.conv2d(x, 32, 5, activation=tf.nn.relu)     ## [batch, y(28), x(28), 1 of channel]
    conv1 = tf.layers.max_pooling2d(conv1, 2, 2)                  ## [batch, y(14), x(14), 1 of channel]
    conv2 = tf.layers.conv2d(conv1, 64, 3, activation=tf.nn.relu) ## [batch, y(14), x(14), 1 of channel] 
    conv2 = tf.layers.max_pooling2d(conv2, 2, 2)                  ## [batch, y(7), x(7), 1 of channel]
    fc1 = tf.contrib.layers.flatten(conv2)                        ## [batch, y(7) * x(7) of channel]
    fc1 = tf.layers.dense(fc1, 1024)                              ## [batch, 1024 of channel]
    fc1 = tf.layers.dropout(fc1, rate=dropout)                    ## [batch, 1024 of channel]
    return tf.layers.dense(fc1, n_classes)                        ## [batch, n_classes of channel]
    
# Placeholders for images and their labels
x = tf.placeholder(tf.float32, shape=[None, 784])                 ## [batch, y(28) * x(28) of channel]
y = tf.placeholder(tf.float32, shape=[None, 10])                  ## [batch, 10 of label]
# Train the Model
train = conv_net(x, 10, dropout=0.25)                             ## [batch, 10 of channel]
loss_op = tf.reduce_mean(tf.nn.softmax_cross_entropy_with_logits(logits=train, labels=y))
optimizer = tf.train.AdamOptimizer(learning_rate=0.001)
batch = mnist.train.next_batch(batch_size=128)
...
# Evaluate the Model
print(session.run([loss_op], feed_dict={x:batch[1], y:batch[0]}))  # BUG: x:batch[0], y:batch[1]
\end{lstlisting}
 \end{minipage}
 \caption{Code having a shape incompatibility (comments are shown for illustration, \toolname does not require them).}
 \label{fig:motivating_code}
\end{figure*}

\fig{fig:API_original} shows the original TensorFlow code for the max-pooling operation (a common operation used in deep learning for computer vision). 
\fig{fig:API_modified} represents the shape-abstracted version of this operation.
In the abstracted version, \toolname obtains the input shape (lines 4-12), modifies it according to the other max-pooling parameters (kernel size, strides, padding) and returns the output shape as an object of type \colorbox{LightCyan1}{\textit{SF\_Operation}}. 
On the other hand, the original TensorFlow API code, though seemingly smaller, involves a function call that is very deeply nested and requires much more computation. 
This reveals the lightweight implementation of \toolname in comparison to TensorFlow.
We believe that using \toolname as a pre-processing step for probing shape incompatibility errors will help programmers gain confidence over their code before extensive training or final deployment of the model.

\begin{figure}[t]
 \centering
\begin{lstlisting}[language=Python]
def max_pool(value, ksize, strides, padding, data_format="NHWC", name=None):
  with ops.name_scope(name, "MaxPool", [value]) as name:
    value = ops.convert_to_tensor(value, name="input")
    return gen_nn_ops.max_pool(
        value,
        ksize=ksize,
        strides=strides,
        padding=padding,
        data_format=data_format,
        name=name)
\end{lstlisting}
 \caption{Original API for max pooling operation.}
 \label{fig:API_original}
\end{figure}

\begin{figure}[t]
 \centering
\begin{lstlisting}[language=Python]
def max_pool(value, ksize, strides, padding, data_format="NHWC", name=None):
  # Currently implemented for "SAME" padding.
  def abst_max_pool(value, ksize, strides):
    if isinstance(value, ops.SF_Operation):
      out_ = value
      while(not isinstance(out_, ops.Tensor)):
        out_ = out_.fwd_func(*out_.input_nodes)
      input_shape = out_.shape
    elif isinstance(value, ops.Tensor):
      input_shape = value.shape
    else:
      raise NotImplementedError
    
    a = (input_shape[1]-ksize[1])/strides[1]+1
    b = (input_shape[2]-ksize[2])/strides[2]+1
    output_shape = [input_shape[0], a, b, input_shape[3]]
    return ops.Tensor(output_shape)

  return ops.SF_Operation([value, ksize, strides], ffnc=abst_max_pool, name="max_pool")
\end{lstlisting}
 \caption{Modified API for max pooling operation.}
 \label{fig:API_modified}
\end{figure}


\section{Approach and Realization} 
\label{sec:core}

This section gives a detailed discussion of how \toolname abstracts shape information for different APIs and how it handles the entire computation.

\subsection{Abstract Domain in \toolname}
\begin{figure}[h!]
 \includegraphics[width=\columnwidth]{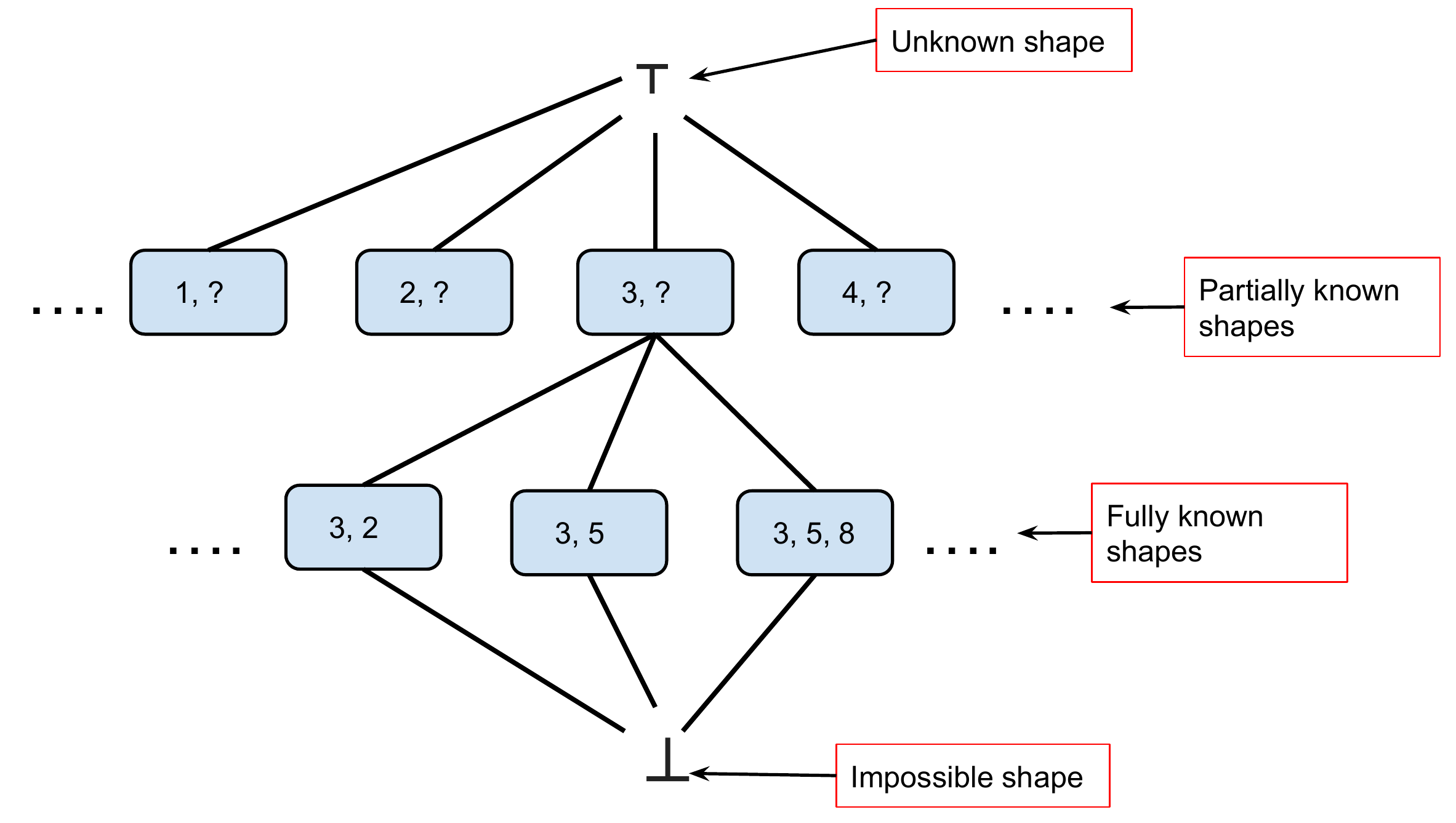}
 \caption{\toolname's abstract domain.}
 \label{fig:lattice}
\end{figure}

This subsection discusses how \toolname abstracts a single operation in TensorFlow code. 
The shapes \toolname's APIs encounters can be broadly categorized into the unknown, partly known, or fully known categories, however, \toolname does not operate on a lattice of these three categories.
\toolname precisely tracks the tensor shapes if it is completely known or partial shape in case of partly known shapes. 
\fig{fig:lattice} shows the abstract domain (which is a lattice) that \toolname uses, and how each of the tensor shapes that it tracks can fall into one of the three categories. 
Note that the lattice in \fig{fig:lattice} is not used for joins because \toolname only performs dynamic analysis. 
Each operation processes its input tensors shapes to generate the output tensor shape, and it is characterized by its specific transfer function. 
For instance, \tab{tab:reshape-transfer-function} shows the transfer function used by an API named ``reshape'', which is intuitively named and is used to change the shape of input tensors. 
It takes as input a tensor and the desired shape and returns the shape of the output tensor (the colored cells of the table). 
A programmer can either fully (e.g. [3, 4, 6]) or partly specify (e.g. [4, -1]) the desired shape. 
The presence of -1 in the desired shape is treated as if the programmer only cares about the dimensions other than the one specified by -1 (partly known desired shape). 
The input tensor itself can be fully known, partly known or be unknown. \toolname tracks the respective shape information. 
Reshape operation allows some kind of inputs (pre-conditions) and returns the output tensor shape if the operation is legal. 
For instance, an unknown desired shape is an illegal input to reshape; therefore, the entire column with ``Unknown'' desired shape is $\bot$ in \tab{tab:reshape-transfer-function}. 
If either the desired shape or input tensor is completely known, the output shape is also completely known, and the precise value of the output shape is returned as a list of integers.

\begin{table}
    \caption{Transfer function for the Reshape operation in \toolname.}
    \label{tab:reshape-transfer-function}
 \begin{tabular}{lc|c|c|c|c}
 \toprule
 \multicolumn{2}{c|}{Reshape Operation} & \multicolumn{4}{c}{Desired shape} \\ \cline{3-6}
 & & Known & Partly known & Unknown & $\bot$ \\ \hline
 \parbox[t]{2mm}{\multirow{4}{*}{\rotatebox[origin=c]{90}{Tensor}}} & \multicolumn{1}{|c|}{Known} & \cellcolor{blue!15}Known & \cellcolor{blue!15} Known & $\bot$ & $\bot$ \\ \cline{2-6}
 & \multicolumn{1}{|c|}{Partly known} & \cellcolor{blue!15}Known & \cellcolor{blue!15}Partly known & $\bot$ & $\bot$ \\ \cline{2-6}
 & \multicolumn{1}{|c|}{Unknown} & \cellcolor{blue!15}Known & \cellcolor{blue!15}Partly known & $\bot$ & $\bot$ \\ \cline{2-6}
 & \multicolumn{1}{|c|}{$\bot$} & $\bot$ & $\bot$ & $\bot$ & $\bot$ \\ 
 \bottomrule
 \end{tabular}
 
\end{table}

\tab{tab:reduce-mean-transfer-function} shows the transfer function for another operation named ``reduce\_mean''. 
It takes as input a tensor and a dimension along which to average the values. 
For simplicity, we consider the case when the programmer specifies no axis, in which case a single integer is supposed to be returned by TensorFlow, which is the average of the entire tensor. 
Since \toolname only tracks the shape of the tensors, irrespective of the input tensor's shape, it always returns one as the output shape of an integer tensor that is 1. 
Note that, 1 is a specific shape in the ``known'' category of shapes, but since ``reduce\_mean'' returns a specific shape from this category, we mention that in the transfer function (\tab{tab:reduce-mean-transfer-function}). 

\begin{table}
 \caption{Transfer function for the Reduce Mean operation in \toolname.}
 \label{tab:reduce-mean-transfer-function}
 \begin{tabular}{l|c|c|c|c}
 \toprule
 Reduce Mean & \multicolumn{4}{c}{Input Tensor1} \\ \cline{2-5}
 Operation & Known & Partly known & Unknown & $\bot$ \\ \hline
 & \cellcolor{blue!15}1 & \cellcolor{blue!15}1 & \cellcolor{blue!15}1 & $\bot$ \\
 \bottomrule
 \end{tabular}
 
\end{table}

\subsection{Computational Graph}
In this subsection, we discuss how \toolname handles a set of operations that need to be executed in a particular sequence. 

The core building blocks in TensorFlow are the computational graphs. 
TensorFlow operates in a lazy paradigm --- it parses the code to first generate a graph which has operations or variables as nodes, and the input and output tensors as directed edges in the graph, and the computation happens when this graph is executed within a session. 
This graph is termed as a computational graph in TensorFlow, and the flow of data in this graph is the basic set of computations in TensorFlow. 
The computational graph encodes the sequence of operations in the Tensorflow code.
\toolname operates on a computational graph which only contains shape information for the input and output tensors for each operation, which we call the \emph{shape computational graph}. 
The structure of the shape computational graph is a replica of the computational graph generated by TensorFlow for the same code but contains no tensor data, only their shapes.

\fig{fig:figure-compute-graph} illustrates a computational graph for the program in \fig{fig:code-compute-graph}, which computes the harmonic mean of the two constants \var{x} and \var{y}.
There are distinct steps in creating the computational graph. 
We first create two constants, \var{x} and \var{y} which have values 11 and 21, respectively. 
We also create a constant \var{z}, which will be used in the calculation of the harmonic mean.
Next, we multiply \var{x} and \var{y} to compute their product (denoted by \var{prod}), which is further multiplied with \var{z} and the outcome is still stored in the variable \var{prod}.
We also add the two constants and store the result in the variable \var{sum}. 
We then divide the two outcomes, \var{prod} and \var{sum}, to return the harmonic mean, save in the variable \var{harmonic}, and finally print it. 
Note that this computational graph is only for illustration purposes and does not contain tensors, which Tensorflow computational graphs have. 

\begin{figure}[t]
\begin{lstlisting}[language=Python]
import tensorflow as tf
x = tf.constant(11, name='x')
y = tf.constant(21, name='y')
z = tf.constant(2, name='z')
prod = tf.multiply(x, y, name="Mul1")
prod = tf.multiply(z, prod, name="Mul2")
sum = tf.add(x, y, name='Add')
harmonic = tf.divide(prod, sum, name="Div")
print(harmonic)
\end{lstlisting}
 \caption{A simple tensorflow program to compute harmonic mean of two constants, x and y.}
 \label{fig:code-compute-graph}
\end{figure}

\begin{figure}[t]
 \includegraphics[width=0.88\columnwidth]{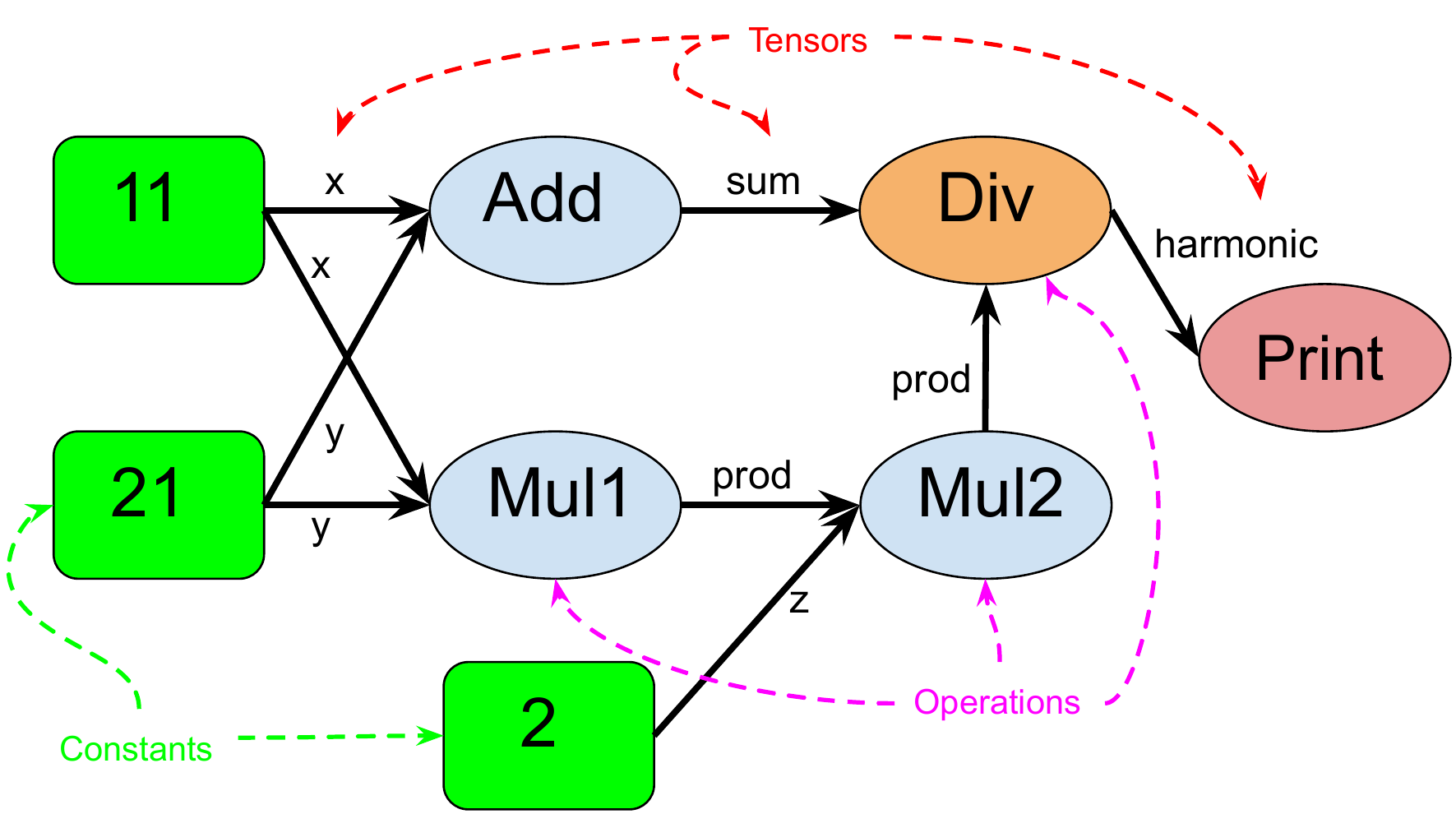}
 \caption{Computational graph of the code to compute harmonic mean of two constants (shown in \fig{fig:code-compute-graph}) with depiction of its data flow.}
 \label{fig:figure-compute-graph}
\end{figure}

A computational graph distinctly specifies the dependencies across different operations, i.e., the data-flow operations. 
Therefore, the computational graph is a directed acyclic graph (DAG), as illustrated by \fig{fig:figure-compute-graph}. 
Highly complex TensorFlow programs can have more than one computational graph, which can be executed in isolation.
When a computational graph representing the code has been constructed, independent operations can be computed in parallel. 
TensorFlow optimizes to distribute parallelizable operations on the same and even across multiple machines.


A typical computational graph consists of several parts:
\begin{itemize}
 \item \emph{Placeholders}: Placeholders can be viewed as the empty nodes in the graph. They represent the empty nodes for data, for example, for training images, that can be instantiated when the graph is to be executed within a session. During graph construction, placeholders require only the shape and type of data that is to be instantiated later.
 \item \emph{Variables}: These are the changeable parameters in a graph.
 \item \emph{Constants}: These hold the non-changeable graph parameters.
 \item \emph{Operations}: These are the nodes that perform computations in the graph. For example, additions and convolutions are operations.
\end{itemize}

A session is runtime execution when the placeholders are populated with data, operations are executed, and the tensors are evaluated in a pre-determined order. 

Since \toolname has modified the implementation of all APIs (which are the nodes in a shape computational graph) to only keep track of shape transformation, the input to any operation node is a shape value, and the output is the transformed shape value (which becomes the input for the child node).
\var{SF\_Operation} is a node in the shape computational graph, on which \toolname operates. 
Each modified function (or API) must return an object of the type \var{SF\_Operation}, and this is added to the shape computational graph at the necessary position, i.e., directed from its input vertex (or vertices) and directed to its output vertex (or vertices).
There are some APIs, e.g., \textit{nn.relu} and \textit{nn.dropout}, which cause no change to the tensor shape; they return the input shape without any computation, one of the reasons that make \toolname significantly faster than TensorFlow. 

The operations in a computational graph must be executed in an order such that all parents of a node must be evaluated before the child node can be evaluated. 
This is ensured by topologically sorting the nodes in a computational graph. 
When a \var{session.run()} function is called, \toolname performs a topological sorting of the shape computational graph.
The \var{session.run()} requires to be called with at least two parameters: an operation (which usually is the last node in a computational graph) and a dictionary which contains concrete values (data) for the placeholders.

Another noticeable difference from TensorFlow is the absence of a backward pass. 
TensorFlow implements a forward and a backward pass for each of its APIs (which feature as vertices in the computational graph). 
This is necessary for optimizations using Stochastic Gradient Descent or any other optimization algorithms. 
For reaching an optimal point, the computational graph needs to be executed many times (depending on the problem, this can be as high as a million times). 
Since \toolname does not need to iterate over the shape computational graph multiple times, it does not have a backward pass method defined for the modified APIs. 

\subsection{\toolname Implementation}
\label{sec:implement}


We have built \toolname inside TensorFlow~\cite{tensorFlow-paper}.
We decide to follow the TensorFlow skeleton because it carries the benefits of being developed by experienced researchers/engineers, and shaped by extensive usage and feedback from the large machine learning community.
By using this strategy, we avoid not only the development of another interpreter (\toolname depends on Python for interpretation), but also the creation of a deeply nested API chain which is shipped by TensorFlow. 
The problem with a nested API is that a programmer can call the same convolution API by either calling \var{tensorflow.nn.conv2d} or \var{nn.conv2d}, or simply \var{conv2d}, depending on how deep TensorFlow imports are used previously in the code. 
Had we built our own interpreter and analyzer, we would be required to follow the exact same nesting as TensorFlow to address this problem. 
We have avoided all these redundancies by constructing \toolname inside TensorFlow.
There are disadvantages associated with this choice --- we had to add some glue code to several APIs in order to support vital parts for TensorFlow, but the advantages outweigh.




Although in principle any TensorFlow API which consumers and emits a tensor can be modified for shape abstraction, in total we modified 118 APIs in TensorFlow. These were the APIs used by the benchmark suite. 
Out of these 118 APIs, 60 APIs were programmer-exposed APIs (i.e., APIs that programmer use in their code) and the rest 58 were internal to the functioning of TensorFlow, but required to be modified in order to build \toolname.
We will make \toolname publicly available on GitHub to benefit TensorFlow users and follow-up research, and allow other researchers/developers to contribute and modify the remaining APIs in order to make \toolname useful for all deep learning code written in TensorFlow.

\section{Evaluation}
\label{sec:evaluate}

This section details our evaluation of \toolname. In particular, 
we evaluate \toolname to answer the following research questions:
\begin{enumerate}
 \item \textbf{RQ1:} How does \toolname compare to prior work in terms of the accuracy of analysis?
 \item \textbf{RQ2:} Is \toolname much faster than TensorFlow in catching shape incompatibility errors?
 \item \textbf{RQ3:} Is \toolname' analysis effective and accurate?
\end{enumerate}

For answering the research questions, we evaluate \toolname on the benchmarks released by Zhang et al.~\cite{issta-bug-collection}, and also on the ones listed by Johirul et al.~\cite{fse-bug-collection}. 
Zhang et al.~\cite{issta-bug-collection} collected 175 buggy TensorFlow programs, 88 programs from Github, and 87 programs from StackOverFlow, respectively. 
They categorized them into the following six principled root causes:
\begin{enumerate}
 \item Incorrect model parameter or structure
 \item Unaligned Tensor or Data Flow/Data bugs (i.e., shape incompatibility errors)
 \item Confused computational model
 \item API change
 \item API misuse
 \item Structural inefficiency
\end{enumerate}
They open-sourced the collection of the buggy programs and also added the fixed versions of the programs. 
We isolate the programs with shape incompatibility errors (labeled as \textit{Unaligned Tensor}) from this collection. 

Johirul et al.~\cite{fse-bug-collection} collected 970 buggy programs written in five deep learning libraries. 
Out of those, 266 programs were in TensorFlow, 100 programs from GitHub, and 166 programs from StackOverflow. 
They categorized the buggy programs based on their bug types, root causes of the bugs, and their effects. 
The categorization of bugs was finer in Johirul et al.'s collection, and hence we combine the programs labeled with \textit{Data Bugs} and \textit{Data Flow bugs} from their suite; they had shape incompatibility errors.
Although they did not provide the code for their programs, they provided the list of the StackOverFlow posts, and GitHub repositories examined. 
For the StackOverflow posts, we used both the buggy and the corresponding correct versions of the programs and will release them with our implementation of \toolname. 
Johirul et al.'s collection of GitHub programs had no shape incompatibility error bugs for the TensorFlow library.
In total, we use 20 StackOverFlow programs from their collection. 



In total, we evaluate \toolname on 52 programs, containing a mix of buggy and correct versions of benchmarks from the bug collection.
For Zhang et al.'s collection, the average lines of code in the StackOverFlow programs is 42 lines, whereas, for the GitHub programs, it is 1,304 lines.
For Johirul et al.'s collection, the average lines of code for the StackOverFlow programs is 36, which is close to the average lines of code in the StackOverFlow benchmarks collected by Zhang et al. 
Their GitHub bug collection records no programs having shape incompatibility errors.
\tab{tab:benchmark-statistics} summarizes these statistics. 

\begin{table}
  \caption{Benchmark statistics.}
 \label{tab:benchmark-statistics}
 \begin{tabular}{ccrr}
 \toprule
 Prior Work & Source & \# programs & Avg. LOC \\
 \midrule
 Zhang et al.~\cite{issta-bug-collection} & StackOverFlow & 26 & 42 \\
 Zhang et al.~\cite{issta-bug-collection} & Github & 6 & 1,304 \\
 Johirul et al.~\cite{fse-bug-collection} & StackOverFlow & 20 & 36 \\
 \bottomrule
 \end{tabular}
 
\end{table}

\subsection{Accuracy of the analysis}
To compare \toolname's analysis to the previous static analysis techniques, we calculate the number of bugs detected by Ariadne and \toolname over the set of benchmark programs. 
Since Ariadne is not open-sourced, we could only compare with it on the set of common benchmarks. 
Ariadne were evaluated on 14 programs which were also selected from Zhang et al.'s collection of StackOverFlow programs. 
\tab{tab:comparison-past-tools} shows the analysis results for Ariadne and \toolname on these 14 benchmarks. 
Ariadne due to its limited support for tensor modifying APIs is not able to report bugs in any of the 14 benchmarks. 
Pythia was also evaluated on the same set of 14 programs and it finds bugs in 11 out of them. 
\toolname sucessfully finds bugs in 13 out of 14 programs. 
The results for Ariadne and Pythia were taken from Table 2 in ~\cite{Pythia-ibm}. 
Therefore, \toolname outperforms Ariadne by 93\% and Pythia by 14\% in terms of the ability to detect bugs.

\begin{table}
  \centering
  \caption{Comparsion of the capability of catching bugs . \toolname detects bugs in more programs than both the other tools. }
  \label{tab:comparison-past-tools}
  \begin{tabular}{ccc}
      \toprule
      Benchmarks    & Ariadne~\cite{Ariadne-ibm}  & \toolname                     \\
      \midrule
      UT-1 buggy    &  \xmark                     &     \checkmark                \\ 
      UT-2 buggy    &  \xmark                     &     \checkmark                \\ 
      UT-3 buggy    &  \xmark                     &     \checkmark                \\ 
      UT-4 buggy    &  \xmark                     &     \checkmark                \\ 
      UT-5 buggy    &  \xmark                     &     \checkmark                \\ 
      UT-6 buggy    &  \xmark                     &     \checkmark                \\ 
      UT-7 buggy    &  \xmark                     &     \xmark                    \\ 
      UT-8 buggy    &  \xmark                     &     \checkmark                \\ 
      UT-9 buggy    &  \xmark                     &     \checkmark                \\ 
      UT-10 buggy   &  \xmark                     &     \checkmark                \\ 
      UT-11 buggy   &  \xmark                     &     \checkmark                \\ 
      UT-12 buggy   &  \xmark                     &     \checkmark                \\ 
      UT-13 buggy   &  \xmark                     &     \checkmark                \\ 
      UT-15 buggy   &  \xmark                     &     \checkmark                \\ 
      \bottomrule
  \end{tabular}
  
\end{table}

\subsection{Experimental Setup and Timing Results}

To answer \textbf{RQ2}, we compare the shape incompatibility error detection time for buggy programs (or \textit{no error detected} for correct programs) with the original TensorFlow. 
For a fair and direct comparison, we compiled TensorFlow and \toolname from the source using the same configuration. 
We also carefully maintained the same versions for TensorFlow and \toolname (TensorFlow version in which \toolname was built into).
We used TensorFlow version 1.8.0.

\begin{table}
  \centering
  \caption{Number of data points in the original and the dummy version of the three datasets our benchmarks used.}
    \label{tab:data-points-datasets}
    \begin{tabular}{crr}
        \toprule
        Dataset         & Data points        & Data points \\
                        & in original dataset &  in dummy dataset  \\
        \midrule
        MNIST           &  60,000  &   2      \\
        CelebA          &  202,599 &   5      \\
        Boston housing  &  506    &   4      \\
        \bottomrule
    \end{tabular}
    
\end{table}

\begin{table*} 
  \caption{\textit{\toolnametable1} and \textit{TensorFlow1} columns respectively record the time taken by \toolname and TensorFlow when run on the full dataset, and \textit{Gain1} column records the speed-up in this case. Similarly, \textit{\toolnametable2} and \textit{TensorFlow2} columns record the respective runtimes when run on a subset of the full dataset, and \textit{Gain2} column records the speed-up obtained in this case.}
    \begin{subtable}{\textwidth}
      \caption{Speed-up obtained by using \toolname for the \emph{StackOverFlow} benchmarks collected by Zhang et. al.~\cite{issta-bug-collection}.}
    \label{tab:issta-sof-results}
      \centering
	  \begin{tabular}{lcc>{\bfseries}acc>{\bfseries}a}
		\toprule
      Benchmarks   &  \toolnametable1      &  TensorFlow1           &  Gain1 &  \toolnametable2      &  TensorFlow2 &  Gain2 \\
                   &  runtime (s)          &  runtime (s)           &        &  runtime (s)          &  runtime (s)  &        \\
      \midrule
       UT-1 buggy  &  0.004826 (0.004555)  &   0.505632 (0.507254)  &  105   &  0.012453 (0.012306)  &  0.210542 (0.210695)   &   17    \\
       UT-1 fix    &  0.011045 (0.011301)  &   2.235000 (2.236960)  &  202   &  0.015355 (0.014678)  &  0.289705 (0.288663)   &   19    \\ 
       UT-2 buggy  &  0.016893 (0.018482)  &   0.020635 (0.020435)  &    1   &  0.020888 (0.023390)  &  0.021937 (0.021691)   &    1    \\
       UT-2 fix    &  0.016432 (0.016418)  &   0.028978 (0.028090)  &    2   &  0.015112 (0.015101)  &  0.028099 (0.028331)   &    2    \\
       UT-3 buggy  &  0.003249 (0.003239)  &   0.005512 (0.005563)  &    2   &  0.003109 (0.003117)  &  0.004920 (0.005415)   &    2    \\
       UT-3 fix    &  0.062063 (0.062306)  &   0.056764 (0.056520)  &    1   &  0.061828 (0.062009)  &  0.057082 (0.057010)   &    1    \\ 
       UT-4 fix    &  0.052681 (0.052392)  &   0.551365 (0.550870   &   10   &  0.050347 (0.050485)  &  0.550860 (0.551190)   &   11    \\
       UT-4 buggy  &  0.016712 (0.018654)  &   0.298007 (0.297025)  &   18   &  0.016412 (0.018195)  &  0.297931 (0.298694)   &   18    \\
       UT-5 fix    &  0.013685 (0.012162)  &   2.156230 (2.164300)  &  158   &  0.020843 (0.020723)  &  1.799380 (1.798460)   &   86    \\
       UT-5 buggy  &  0.010792 (0.011070)  &   1.546890 (1.547420)  &  143   &  0.016171 (0.015626)  &  1.268940 (1.264490)   &   78    \\
       UT-6 buggy  &  0.016913 (0.016393)  &   0.029426 (0.029343)  &    2   &  0.020511 (0.023374)  &  0.026882 (0.027900)   &    1    \\
	     UT-6 fix    &  0.016028 (0.014938)  &   0.181201 (0.182266)  &   11   &  0.013293 (0.014880)  &  0.182537 (0.182204)   &   14    \\
       UT-8 buggy  &  0.004550 (0.004782)  &   0.011710 (0.011202)  &    3   &  0.004267 (0.004791)  &  0.010029 (0.010495)   &    2    \\
       UT-8 fix    &  0.014360 (0.014119)  &   0.234234 (0.099028)  &   16   &  0.017531 (0.016500)  &  0.098577 (0.098470)   &    6    \\
       UT-9 fix    &  0.018558 (0.018542)  &   0.207446 (0.142824)  &   11   &  0.016070 (0.014876)  &  0.138113 (0.139519)   &    9    \\
       UT-9 buggy  &  0.019071 (0.017793)  &   0.210779 (0.142963)  &   11   &  0.020065 (0.019071)  &  0.142899 (0.142407)   &    7    \\
       UT-10 buggy &  0.476457 (0.473398)  &   0.535116 (0.529867)  &    1   &  0.016264 (0.017632)  &  0.176798 (0.098980)   &   11    \\
       UT-10 fix   &  0.722202 (0.723635)  &  24.845400 (24.83100)  &   34   &  0.017035 (0.015083)  &  0.231835 (0.156492)   &   14    \\
       UT-11 buggy &  0.082495 (0.084547)  &   0.090225 (0.090912)  &    1   &  0.087516 (0.088000)  &  0.092765 (0.092045)   &    1    \\
       UT-11 fix   &  0.053065 (0.057801)  &   0.067534 (0.069481)  &    1   &  0.055528 (0.054685)  &  0.064511 (0.064389)   &    1    \\
       UT-12 buggy &  0.021083 (0.019338)  &   0.416095 (0.333569)  &   20   &  0.166605 (0.168636)  &  0.646697 (0.478289)   &    4    \\
       UT-12 fix   &  0.012945 (0.013822)  &   0.356951 (0.355827)  &   28   &  0.162853 (0.162241)  &  1.116350 (1.051680)   &    7    \\
       UT-13 fix   &  0.017249 (0.016833)  &   0.311994 (0.170065)  &   18   &  0.019410 (0.020330)  &  0.264122 (0.173946)   &   14    \\
       UT-13 buggy &  0.014732 (0.016925)  &   0.173761 (0.173950)  &   12   &  0.016721 (0.014648)  &  0.246741 (0.170383)   &   15    \\
       UT-15 fix   &  0.015963 (0.014282)  &   0.134329 (0.133624)  &    8   &  0.015457 (0.016975)  &  0.266368 (0.135218)   &   17    \\
       UT-15 buggy &  0.016604 (0.015436)  &   0.134649 (0.135490)  &    8   &  0.014803 (0.014519)  &  0.135418 (0.134586)   &    9    \\
	   \midrule  
	Overall Speedup  &             &              &   32   &             &              &   14 \\
      \bottomrule
      \end{tabular}
    
    \end{subtable}%
    
    \begin{subtable}{\textwidth}
      \caption{Speed-up obtained by using \toolname for the \emph{Github} benchmarks collected by Zhang et. al.~\cite{issta-bug-collection}.}
      \label{tab:issta-github-results}
      \centering
	  \begin{tabular}{lcc>{\bfseries}acc>{\bfseries}a}
		\toprule
      Benchmarks       & \toolnametable1         &  TensorFlow1          & Gain1   & \toolnametable2       &  TensorFlow2          &  Gain2     \\
                       &  runtime (s)            & runtime (s)           &         & runtime (s)           & runtime (s)           &        \\
      \midrule    
       UT-1 buggy      &    0.013490 (0.013551)  &   2.72417 (2.41282)   &   202   &  0.019120 (0.018631)  &  0.256937 (0.255222)  &  13    \\
       UT-1 fix        &    0.007925 (0.007502)  &   2.04975 (2.02793)   &   259   &  0.014209 (0.014493)  &  0.238498 (0.236754)  &  17    \\
       UT-7 buggy      &    0.029057 (0.029233)  &   5.10912 (5.03372)   &   176   &  0.038649 (0.036645)  &  2.443070 (2.443690)  &  63    \\
	     UT-7 fix        &    0.022813 (0.023250)  &   536.835 (535.343)   & 23,532   &  0.015904 (0.016015)  &  0.627817 (0.631867)  &  39    \\
       UT-9 buggy      &    0.811806 (0.795560)  &   3.03758 (3.05285)   &     4   &  0.085000 (0.086116)  &  2.415270 (2.406110)  &  28    \\
       UT-9 fix        &  554.639000 (549.1520)  &   Timeout             &   N/A   &  0.066380 (0.066954)  &  2.471560 (2.465310)  &  37    \\
	   \midrule  
   Overall Speedup     &               &              &  4,835   &             &              &  33    \\	   
      \bottomrule
      \end{tabular}
      
    \end{subtable}

\end{table*}

For a realistic comparison, we measure \toolname's performance with two baselines.
In deep learning, practitioners often create a smaller dataset to test parts of their code.
After they are satisfied with results, the code is used on the full dataset for training a model.
The two baselines that we consider differ in the size of the dataset, which needs to be pre-processed and loaded before the computational graph (shape computational graph in case of \toolname) is executed.
In the first baseline, we measure the gain obtained by using \toolname over TensorFlow for the case where the programmer directly executes the code over the full dataset.
For the second baseline, we assume the programmer has created a dummy dataset for testing the code.
We manually created these dummy datasets for the original datasets required by all our benchmarks, which were MNIST~\cite{lecun-mnisthandwrittendigit-2010}, CelebA~\cite{celebA_dataset}, and the Boston housing data~\cite{boston_housing}. 
\tab{tab:data-points-datasets} records the number of data points in the original and dummy versions of these three datasets.

To measure time, we use \var{time} module~\cite{python_reference} of Python because it has a much higher precision that the Unix \var{time} utility~\cite{unix_utilities}. 
In order to amortize runtime variability, all the programs are executed five times. 
We record the average and medians (in parentheses) of the runtimes.
We report timings (in seconds) up to 6 digits after the decimal place.

We ran all the benchmarks for only one epoch. Our benchmarks do not have program paths that are executed only when $epoch > 1$. 
Also, since our benchmarks contain several correct programs, if we run the code for more than one epoch, it would unfairly bias the results against TensorFlow.
Note that we run for all mini-batches of training data in the first (and only) epoch, the reason for which is discussed next.
\fig{fig:second-minibatch} shows a program that throws an error only from the second mini-batch onwards. 
Lines 2 and 3 are the shapes of the variables \var{input} and \var{store}.
After performing the matrix multiplication (line 5), the result is stored in \var{store} after its transposition. 
The error occurs on line 5 in the second minibatch when it attempts to multiply two matrices with incompatible shapes. 
It is not difficult to conceive a program that can give error on the last minibatch.
This was the reason for choosing an experimental design that takes all-mini-batches in the first epoch.

\begin{figure}
\begin{lstlisting}[language=Python]
batches = 5
input_shape = [3, 6]
initial_store_shape = [6, 4]
for mini_batches in range(batches):
  store = input * store  # (matrix multiply, resulting shape = [3, 4])
  store = store.T        # (transpose, resulting shape = [4, 3])
\end{lstlisting}
	\caption{Example of a TensorFlow program which throws an error only after first mini-batch.}
	\label{fig:second-minibatch}
\end{figure}


\begin{figure}
\begin{lstlisting}[language=Python]
import tensorflow as tf, numpy as np

x = tf.placeholder(tf.float32, [None])
x.set_shape([1028178])
y = tf.identity(x)
y.set_shape([478, 717, 3]) #Bug: Can't reset shape
X = np.random.normal(0, 0.1, 1028178)

sess = tf.Session()
Y = sess.run(y, feed_dict={x: X})
\end{lstlisting}
	\caption{Example of a TensorFlow program with small computational graph.}
	\label{fig:tiny-code}
\end{figure}

\begin{table*}
  \caption{Speed-up gained for programs taken from the \emph{StackOverFlow} posts collected by Johirul et al.~\cite{fse-bug-collection}. \textit{\toolnametable1} and \textit{TensorFlow1} columns respectively record the time taken by \toolname and TensorFlow when run on the full dataset, and \textit{Gain1} records the speed-up in this case. \textit{\toolnametable2}, \textit{TensorFlow2}, and \textit{Gain2} record the respective runtimes and speed-up when run on a subset of the full dataset.}
    \label{tab:fse19-results}
  \centering
    \begin{tabular}{lcc>{\bfseries}acc>{\bfseries}a}
      \toprule
	  Benchmarks           &  \toolnametable1      &  TensorFlow1          &  Gain1  &  \toolnametable2     &  TensorFlow2          &  Gain2       \\
                         &  runtime (s)          &  runtime (s)          &         &  runtime (s)         &  runtime (s)          &        \\    
  \midrule
       340892850 buggy   & 0.004327  (0.004170)  & 0.032657 (0.030197)   &      8  & 0.003728 (0.003471)  & 0.031075 (0.031325)   &    8        \\
       340892850 correct & 0.000899  (0.000693)  & 0.039737 (0.039529)   &     44  & 0.000628 (0.000581)  & 0.042388 (0.041677)   &   68        \\
       38399609 buggy    & 0.067773  (0.068316)  & 0.341563 (0.341291)   &      5  & 0.008441 (0.008277)  & 0.341541 (0.340354)   &   40        \\
       38399609 correct  & 0.058631  (0.058109)  & 0.336807 (0.338689)   &      6  & 0.002969 (0.002957)  & 0.335843 (0.335065)   &  113        \\
       37444951 buggy    & 0.004292  (0.004238)  & 0.064222 (0.060475)   &     15  & 0.003693 (0.003587)  & 0.062044 (0.061669)   &   17        \\
       37444951 correct  & 0.000840  (0.000800)  & 0.070589 (0.074351)   &     84  & 0.000724 (0.000606)  & 0.074444 (0.072752)   &  103        \\
       39032277 buggy    & 0.022443  (0.021815)  & 0.892349 (0.892292)   &     40  & 0.020334 (0.020277)  & 0.882298 (0.879868)   &   43        \\
       39032277 correct  & 0.018372  (0.017770)  & 0.894322 (0.887674)   &     49  & 0.018448 (0.018779)  & 0.887020 (0.884134)   &   48        \\
       35488717 buggy    & 0.022591  (0.023691)  & 0.032989 (0.032877)   &      1  & 0.019610 (0.019100)  & 0.033474 (0.033891)   &    2        \\
       35488717 correct  & 0.016953  (0.016514)  & 0.025879 (0.025022)   &      2  & 0.016024 (0.016271)  & 0.028250 (0.025550)   &    2        \\
       36870792 buggy    & 0.003583  (0.003514)  & 0.030363 (0.029801)   &      8  & 0.003645 (0.003618)  & 0.029981 (0.029758)   &    8        \\
       36870792 correct  & 0.000185  (0.000192)  & 0.028432 (0.029228)   &    154  & 0.000170 (0.000163)  & 0.030539 (0.030009)   &  180        \\
       40430186 correct  & 0.060506  (0.059769)  & 0.385061 (0.229717)   &      6  & 0.066676 (0.067285)  & 0.284475 (0.232514)   &    4        \\
       40430186 buggy    & 0.061721  (0.064051)  & 0.219258 (0.215891)   &      4  & 0.057335 (0.057502)  & 0.220841 (0.218171)   &    4        \\
       34908033 buggy    & 0.003358  (0.003445)  & 0.025395 (0.027033)   &      8  & 0.003204 (0.003216)  & 0.026353 (0.025084)   &    8        \\
       34908033 correct  & 0.016301  (0.016568)  & 0.272701 (0.115837)   &     17  & 0.015918 (0.016526)  & 0.264848 (0.114233)   &   17        \\
       38114534 buggy    & 0.019703  (0.017871)  & 0.030618 (0.030867)   &      2  & 0.018022 (0.016896)  & 0.024843 (0.025435)   &    1        \\
       38114534 correct  & 0.016325  (0.016593)  & 0.022600 (0.025680)   &      1  & 0.016263 (0.016469)  & 0.025113 (0.023905)   &    2        \\
       47308181 buggy    & 0.001928  (0.001950)  & 0.002032 (0.002044)   &      1  & 0.001858 (0.001802)  & 0.002031 (0.001887)   &    1        \\
	     47308181 correct  & 0.000593  (0.000569)  & 0.000617 (0.000608)   &      1  & 0.000555 (0.000574)  & 0.000572 (0.000532)   &    1        \\
	   \midrule
	   Overall Speedup     &             &              &     23 &             &              &     34         \\	   
      \bottomrule
    \end{tabular}

\end{table*}

We adhere to the following taxonomy in \tab{tab:issta-sof-results}, \tab{tab:issta-github-results} and \tab{tab:fse19-results}:
columns labelled as \toolnametable1 and TensorFlow1 record the time taken by \toolname and TensorFlow respectively for the first baseline. 
Values in column Gain1 are obtained by dividing respective rows in \toolnametable1 and TensorFlow1 and shows the speed-up \toolname achieves. 
Similarly, \toolnametable2 and TensorFlow2 columns record the time taken by \toolname and TensorFlow respectively for the second baseline. 
Gain2 registers the speed-up procured for this baseline.


\tab{tab:issta-sof-results} presents the timing and speed-up results for the StackOverFlow benchmarks collected by Zhang et al. 
The first three columns deal with the first baseline, and the last three columns deal with the second baseline.
The overall average speed-up for the first baseline is 32X, and for the second baseline is 14X. 
The relative variance for all the benchmarks was smaller than 0.5\%. 

\tab{tab:issta-github-results} presents results for Github programs from the same bug collection by Zhang et al. 
Noticeably, one of the benchmarks times-out (1 hour) for TensorFlow, but \toolname was able to analyze it.
In this case, the overall average speed-up for the first baseline is 4,835X, and for the second baseline is 33X. 
The relative variance for all the benchmarks, but \textit{UT-9 fix} was smaller than 0.5\%. 
This shows that larger programs (GitHub programs were much larger than StackOverFlow programs on average) are expected to gain more from using \toolname. 
The \textit{UT-9} benchmark uses CelebA dataset which consists of more than 200K images. 
Therefore, even \toolname takes significant time (9-10 minutes) to analyse the fixed version of the program, but TensorFlow times-out after a long wait.

This also explains the reason behind several rows in \tab{tab:issta-sof-results} where the gain by using \toolname is insignificant (Gain1 and Gain2 values equalling 1 or 2). 
If the computational graph is very small, the user does not gain much by using \toolname.
\fig{fig:tiny-code} shows such a program, in which only two placeholders were created. 
There is a bug on line 6, where the user unknowingly sets the shape of a \var{placeholder}, which has already been set on line 4. 
\toolname and TensorFlow take almost equal time to reach the error.


\tab{tab:fse19-results} presents the timing and speed-up results for the programs collected from Johirul et al.~\cite{fse-bug-collection}. 
The column nomenclature remains unchanged.
For these programs, the overall average speed-up for the first baseline is 23X, and for the second baseline is 34X. 
The relative variance for all the benchmarks was smaller than 0.5\%. 
We note that these gains are similar to the ones shown in~\tab{tab:issta-sof-results}, and
this corresponds well with the fact that they have similar program length.

Overall, across the 52 benchmarks, \toolname's average speed-up for the first baseline is 499X, and for the second baseline is 24X. 
Therefore, for correct programs the overhead of running \toolname is about 0.2\% in the case of first baseline and 4\% in the case of second baseline. 
On the other hand, for buggy programs \toolname is able to catch bugs within this highly reduced time frame, much quicker than TensorFlow. 

Our purpose to include the correct programs in our suite of benchmarks is two-fold. 
First, it shows that if \toolname is used on correct programs (as the user's code may be correct and not have shape incompatibility errors), it is still fast and offers drastically reduced runtime over the vanilla TensorFlow. 
Second, by including the fixed code, we show that \toolname is not only fast, but also accurate and does not raise false alarms on the fixed code. 


\subsection{Correctness of Implementation}
To answer \textbf{RQ3}, we examined the number of false positives (correct program being labeled as buggy) and false negatives (buggy programs flagged as having no errors).
From the benchmarks that we collected, \toolname generated one false negative. 
We have removed the runtime performance for that program from our experimental results.
Upon further investigation, we found that the bug being thrown by TensorFlow was, in fact, an API misuse flag, rather than a shape incompatibility error, but to be consistent with respect to the classification used by Zhang et al., we consider this program a false negative
for \toolname.\footnote{The program label is \var{UT-7} in the bug collection of~\cite{issta-bug-collection}}.




\subsection{Threats to validity}
We show the accuracy of the analysis of \toolname and its timing performance on an extensive set of benchmarks. 
To eliminate bias, the benchmarks we selected were collected from other studies.
Nevertheless, \toolname can be evaluated on yet larger and challenging benchmarks, which makes interesting future work. 


\vspace{-0.2em}
\section{Related Work}
\label{sec:related}

\pg{Bug detection in machine learning code}
Although there have been attempts in improving visualization of computational graphs in machine learning, to help interpretability and debugging of models~\cite{visualize1, visualize2, visualize3}, there has not been much focus on automated or faster deep-learning code analysis.

Ariadne~\cite{Ariadne-ibm} is a static analysis tool for machine learning code written in TensorFlow.
It uses the existing program analysis framework WALA~\cite{wala-page} for analyzing deep learning code.
It converts Python code (the language in which most deep learning code is written) into WALA's Internal Representation (IR) for analysis.
The programmer must annotate each line of the code in which any tensor operation takes place to indicate the expected output shape after the execution of that operation.
Once the programmer annotates the code with information about the shape of the input data, Ariadne uses dataflow analysis in a tensor tracking type system to verify whether the expected shapes are satisfied at each annotated line.
Ariadne was only evaluated on whether it verified correct code written in TensorFlow, so we are unaware of how it behaves with buggy code.
Given its preliminary implementation (4 APIs only), programmers cannot use Ariadne for most
nontrivial TensorFlow code. 
We have discussed earlier (\sect{sec:intro}) how \toolname and Ariadne differ. 
Pythia, the recent extension of Ariadne is also a static analysis tool and therefore had to develop a whole Python front-end (parser, IR generator) that translates the Python source code into the IR of the WALA framework~\cite{wala-page}, 
a generator of relational tables for declarative program analysis in the Doop framework~\cite{doop-framework}, 
and a points-to, constant-flow and call-graph analysis for Python.
Pythia is a concurrent work to ours and we show that \toolname has a more accurate analysis compared to even Pythia.

There has been work in typesafe abstraction for tensor using Scala~\cite{scala-tensor}. 
Authors argue that this could form the basis of future typesafe deep-learning frameworks that are built on Scala. 
Language incompatibility would not allow this to help the problem \toolname addresses. 

There is work that implements deep-learning framework in a typesafe language like Java~\cite{java-deep} and Haskell~\cite{haskell-deep}; however, they are in preliminary stages of development and have yet to gain popularity.

\pg{Empirical studies on machine learning bugs}
Zhang et al.~\cite{issta-bug-collection} collected 175 TensorFlow bugs, from StackOverflow and GitHub. 
They categorized them by their root causes and effects, and common bug fixing patterns.
They discussed the problems faced by deep-learning programmers, such as bug localization.
Johirul et al.~\cite{fse-bug-collection} investigated 2,716 posts from StackOverFlow, and 500 commits from GitHub to isolate bugs in five deep-learning libraries: TensorFlow, Pytorch, Keras, Caffe, and Theano.
This work also categorized the bugs based on their root causes and their impacts along with giving insights into their frequency.
They also discuss several anti-patterns in bugs commonly found in the deep-learning code.


\pg{Machine learning testing}
Extensive work exists on testing the otherwise ``non-testable" domain of machine learning programs. 
When conducting ML testing, all its tightly intertwined components, including data, learning program (SVM, neural networks, etc.) and framework (TensorFlow, Pytorch, etc.), need to be examined for bugs.
Breck et al.~\cite{breck:2019} proposed a validation algorithm for incoming data. 
The system is used on a large scale at Google.
Krishnan et al. proposed BoostClean~\cite{boostclean}, which can be used to detect violations in domain values in training data automatically.
Pham et al.~\cite{cradle} proposed Cradle for testing machine learning frameworks/libraries. 
Xiao et al.~\cite{security-problem} worked on security vulnerabilities in popular deep learning frameworks like TensorFlow, Caffe, and Pytorch.

Metamorphic testing uses multiple implementations of semantically equivalent programs and compares their output on different inputs. 
Murphy et al.~\cite{Murphy_anapproach} proposed the first few metamorphic relations applicable to the machine learning domain. Since then, they have been adapted to be applied to various problems. 
Xie et al.~\cite{Xie:2011} proposed specific metamorphic relations for testing supervised classifiers, which were able to reveal 39 out of the 43 injected bugs in the Weka~\cite{weka} implementations of those classifiers.
Zhou et al.~\cite{Zhou2018MetamorphicTF} proposed MT4MT, which used metamorphic relations to evaluate the translation consistency of machine translation systems.

Other software testing inspired techniques being applied to the machine learning domain include mutation testing~\cite{DeepMutationMT}, combinatorial testing~\cite{deepct}, and fuzzing~\cite{tensorfuzz}.

There also is a focus on testing machine learning models for non-functional but desired properties, like robustness, fairness, and interpretability.
DeepFool~\cite{deepfool} was proposed by Moosavi-Dezfooli et al. to compute perturbations that \textit{fool} the deep neural networks to quantify their robustness.
There is widespread adoption of adversarial input generation~\cite{deeproad, deepguage, adversarial2, surprise} to test the robustness of machine learning systems for applications like autonomous driving.
DeepTest~\cite{deeptest} was designed to test CNNs and RNNs. 
It used a template with nine image transformations to generate realistic data.
DeepRoad~\cite{deeproad} used GANs to generate authentic looking images for autonomous driving. 
Udacity Challenge Dataset images were used as seed input. 
DeepHunter~\cite{deephunter} uses a metamorphic testing strategy to detect defects in neural networks. 
DeepStellar~\cite{deepstellar} uses adversarial techniques for testing recurrent neural networks (RNNs). 
Dwarakanath et al.~\cite{meta-accenture, meta-accenture2} developed metamorphic tests specifically for testing image classifiers and deep learning forecasters. 
DeepXplore~\cite{deepxplore} tests machine learning systems using a white-box differential testing technique to generate test inputs. 
Inspired by the criterion for line and condition coverage in traditional software, they proposed neuron-coverage as a parallel in the machine learning domain. 

Research in fairness in machine learning focuses on discovering, comprehending, and mitigating causes of bias in models.
There are multiple definitions of fairness in the literature with no firm consensus~\cite{fairness-definitions}. 
Galhotra et al. proposed Themis~\cite{themis}, which uses a random strategy for generating test cases for a biased learned model based on the definition of individual fairness~\cite{individual-fair}. 
These test cases are termed as discriminating.
Aequitas~\cite{aequitas} uses a directed testing strategy for a faster and effective generation of discriminating test cases.
Doshi-Velez et al.~\cite{DoshiVelez2017} gave terminologies for the interpretability of machine learning models.
Chen et al.~\cite{Chen2018Calibration} examined several methods for improving interpretability for classifiers.

We refer the reader to a recent survey of the papers at the intersection of software engineering and deep learning~\cite{ferreira2019software}.

\pg{Fixing Machine Learning Bugs}
Due to the unconfined (and sometimes undefined) nature of the bugs in machine learning, fixing them is nontrivial. 
Data resampling has emerged as one of the leading approaches in fixing models. 
The generated test inputs from tools like DeepXplore~\cite{deepxplore} and DeepTest~\cite{deeptest} can be added to the training set to improve accuracy. DeepTest improved the model's accuracy by 46\%.
Ma et al. proposed MODE~\cite{mode} as a tool for identifying \textit{faulty neurons}, neurons that are responsible for misclassification, and for separating training data that influenced such neurons. 
It helped improve model performance by a significant margin.
Vartak et al~\cite{Vartak:2018} proposed the MISTIQUE system, which stores and queries model intermediates to help debug.

We refer the reader to Zhang et al.'s recent survey~\cite{mlt-survey} for a thorough discussion of the work in machine learning testing and associated areas.
All existing efforts surveyed above are orthogonal to \toolname --- none tackles the detection of bugs in machine learning code, which is the focus of this work. 


\section{Conclusion}
\label{sec:conclusion}

We have designed and developed a dynamic shape abstract interpreter
for TensorFlow programs with the following goals:

\begin{enumerate}
  \item The analysis should be light and fast; 
  \item The analysis should be accurate; and 
  \item The analysis should not cause programmer burden.
\end{enumerate}

We have evaluated \toolname on 52 benchmarks and demonstrated its effectiveness in the detection of shape incompatibility errors, which should help save valuable developer time.
Our current implementation modifies 118 APIs out of the thousands in TensorFlow.
We open-source \toolname to benefit both the user and research communities, and to
allow contributions from the wider open-source community to refine \toolname and support
additional APIs and functionalities. 
Interesting future work would include the generation of useful error
messages, localization of source code generating shape
incompatibility errors, and automatic repair of such errors.



\bibliographystyle{ACM-Reference-Format}
\balance
\bibliography{bibfile.bib}


\begin{thebibliography}{57}


\ifx \showCODEN    \undefined \def \showCODEN     #1{\unskip}     \fi
\ifx \showDOI      \undefined \def \showDOI       #1{#1}\fi
\ifx \showISBNx    \undefined \def \showISBNx     #1{\unskip}     \fi
\ifx \showISBNxiii \undefined \def \showISBNxiii  #1{\unskip}     \fi
\ifx \showISSN     \undefined \def \showISSN      #1{\unskip}     \fi
\ifx \showLCCN     \undefined \def \showLCCN      #1{\unskip}     \fi
\ifx \shownote     \undefined \def \shownote      #1{#1}          \fi
\ifx \showarticletitle \undefined \def \showarticletitle #1{#1}   \fi
\ifx \showURL      \undefined \def \showURL       {\relax}        \fi
\providecommand\bibfield[2]{#2}
\providecommand\bibinfo[2]{#2}
\providecommand\natexlab[1]{#1}
\providecommand\showeprint[2][]{arXiv:#2}

\bibitem[\protect\citeauthoryear{??}{bos}{2019}]%
        {boston_housing}
 \bibinfo{year}{2019}\natexlab{}.
\newblock \bibinfo{title}{Boston Housing Dataset}.
\newblock
  \bibinfo{howpublished}{\url{https://www.cs.toronto.edu/~delve/data/boston/bostonDetail.html}}.
\newblock
\newblock
\shownote{Accessed: 2019-10-30.}


\bibitem[\protect\citeauthoryear{??}{tf-}{2019a}]%
        {tf-popular1}
 \bibinfo{year}{2019}\natexlab{a}.
\newblock \bibinfo{title}{Deep Learning Framework Scores}.
\newblock
  \bibinfo{howpublished}{\url{https://towardsdatascience.com/deep-learning-framework-power-scores-2018-23607ddf297a}}.
\newblock
\newblock
\shownote{Accessed: 2019-10-30.}


\bibitem[\protect\citeauthoryear{??}{tf-}{2019b}]%
        {tf-popular3}
 \bibinfo{year}{2019}\natexlab{b}.
\newblock \bibinfo{title}{Growth of Deep Learning Framework}.
\newblock
  \bibinfo{howpublished}{\url{https://www.kdnuggets.com/2019/05/which-deep-learning-framework-growing-fastest.html}}.
\newblock
\newblock
\shownote{Accessed: 2019-10-30.}


\bibitem[\protect\citeauthoryear{??}{mni}{2019}]%
        {mnist-tutorial}
 \bibinfo{year}{2019}\natexlab{}.
\newblock \bibinfo{title}{MNIST Tutorial for TensorFlow}.
\newblock
  \bibinfo{howpublished}{\url{https://github.com/aymericdamien/TensorFlow-Examples/blob/master/examples/3_NeuralNetworks/convolutional_network.py}}.
\newblock
\newblock
\shownote{Accessed: 2019-10-30.}


\bibitem[\protect\citeauthoryear{??}{tf-}{2019c}]%
        {tf-popular2}
 \bibinfo{year}{2019}\natexlab{c}.
\newblock \bibinfo{title}{Top 5 Deep Learning Frameworks for 2019}.
\newblock
  \bibinfo{howpublished}{\url{https://www.springboard.com/blog/deep-learning-frameworks/}}.
\newblock
\newblock
\shownote{Accessed: 2019-10-30.}


\bibitem[\protect\citeauthoryear{??}{jav}{2020}]%
        {java-deep}
 \bibinfo{year}{2020}\natexlab{}.
\newblock \bibinfo{title}{Deep Learning for Java}.
\newblock \bibinfo{howpublished}{\url{https://deeplearning4j.org}}.
\newblock
\newblock
\shownote{Accessed: 2020-2-28.}


\bibitem[\protect\citeauthoryear{??}{has}{2020}]%
        {haskell-deep}
 \bibinfo{year}{2020}\natexlab{}.
\newblock \bibinfo{title}{Deep Learning in Haskell}.
\newblock \bibinfo{howpublished}{\url{https://github.com/HuwCampbell/grenade}}.
\newblock
\newblock
\shownote{Accessed: 2020-2-28.}


\bibitem[\protect\citeauthoryear{Abadi, Barham, Chen, Chen, Davis, Dean, Devin,
  Ghemawat, Irving, Isard, Kudlur, Levenberg, Monga, Moore, Murray, Steiner,
  Tucker, Vasudevan, Warden, Wicke, Yu, and Zheng}{Abadi et~al\mbox{.}}{2016}]%
        {tensorFlow-paper}
\bibfield{author}{\bibinfo{person}{Martin Abadi}, \bibinfo{person}{Paul
  Barham}, \bibinfo{person}{Jianmin Chen}, \bibinfo{person}{Zhifeng Chen},
  \bibinfo{person}{Andy Davis}, \bibinfo{person}{Jeffrey Dean},
  \bibinfo{person}{Matthieu Devin}, \bibinfo{person}{Sanjay Ghemawat},
  \bibinfo{person}{Geoffrey Irving}, \bibinfo{person}{Michael Isard},
  \bibinfo{person}{Manjunath Kudlur}, \bibinfo{person}{Josh Levenberg},
  \bibinfo{person}{Rajat Monga}, \bibinfo{person}{Sherry Moore},
  \bibinfo{person}{Derek~G. Murray}, \bibinfo{person}{Benoit Steiner},
  \bibinfo{person}{Paul Tucker}, \bibinfo{person}{Vijay Vasudevan},
  \bibinfo{person}{Pete Warden}, \bibinfo{person}{Martin Wicke},
  \bibinfo{person}{Yuan Yu}, {and} \bibinfo{person}{Xiaoqiang Zheng}.}
  \bibinfo{year}{2016}\natexlab{}.
\newblock \showarticletitle{TensorFlow: A system for large-scale machine
  learning}. In \bibinfo{booktitle}{\emph{12th USENIX Symposium on Operating
  Systems Design and Implementation (OSDI 16)}}. \bibinfo{pages}{265--283}.
\newblock
\urldef\tempurl%
\url{https://www.usenix.org/system/files/conference/osdi16/osdi16-abadi.pdf}
\showURL{%
\tempurl}


\bibitem[\protect\citeauthoryear{Bravenboer and Smaragdakis}{Bravenboer and
  Smaragdakis}{2009}]%
        {doop-framework}
\bibfield{author}{\bibinfo{person}{Martin Bravenboer} {and}
  \bibinfo{person}{Yannis Smaragdakis}.} \bibinfo{year}{2009}\natexlab{}.
\newblock \showarticletitle{Strictly Declarative Specification of Sophisticated
  Points-to Analyses}. In \bibinfo{booktitle}{\emph{Proceedings of the 24th ACM
  SIGPLAN Conference on Object Oriented Programming Systems Languages and
  Applications}} \emph{(\bibinfo{series}{OOPSLA '09})}.
  \bibinfo{publisher}{Association for Computing Machinery},
  \bibinfo{address}{New York, NY, USA}, \bibinfo{pages}{243–262}.
\newblock
\showISBNx{9781605587660}
\urldef\tempurl%
\url{https://doi.org/10.1145/1640089.1640108}
\showDOI{\tempurl}


\bibitem[\protect\citeauthoryear{Breck, Zinkevich, Polyzotis, Whang, and
  Roy}{Breck et~al\mbox{.}}{2019}]%
        {breck:2019}
\bibfield{author}{\bibinfo{person}{Eric Breck}, \bibinfo{person}{Marty
  Zinkevich}, \bibinfo{person}{Neoklis Polyzotis}, \bibinfo{person}{Steven
  Whang}, {and} \bibinfo{person}{Sudip Roy}.} \bibinfo{year}{2019}\natexlab{}.
\newblock \showarticletitle{Data Validation for Machine Learning}. In
  \bibinfo{booktitle}{\emph{Proceedings of SysML}}.
\newblock


\bibitem[\protect\citeauthoryear{Cai, Breck, Nielsen, Salib, and Sculley}{Cai
  et~al\mbox{.}}{2016}]%
        {visualize2}
\bibfield{author}{\bibinfo{person}{Shanqing Cai}, \bibinfo{person}{Eric Breck},
  \bibinfo{person}{Eric Nielsen}, \bibinfo{person}{Michael Salib}, {and}
  \bibinfo{person}{D. Sculley}.} \bibinfo{year}{2016}\natexlab{}.
\newblock \showarticletitle{TensorFlow Debugger: Debugging Dataflow Graphs for
  Machine Learning}.
\newblock


\bibitem[\protect\citeauthoryear{Chan, Ma, Juefei{-}Xu, Xie, Liu, and Ong}{Chan
  et~al\mbox{.}}{2018}]%
        {adversarial2}
\bibfield{author}{\bibinfo{person}{Alvin Chan}, \bibinfo{person}{Lei Ma},
  \bibinfo{person}{Felix Juefei{-}Xu}, \bibinfo{person}{Xiaofei Xie},
  \bibinfo{person}{Yang Liu}, {and} \bibinfo{person}{Yew~Soon Ong}.}
  \bibinfo{year}{2018}\natexlab{}.
\newblock \showarticletitle{Metamorphic Relation Based Adversarial Attacks on
  Differentiable Neural Computer}.
\newblock \bibinfo{journal}{\emph{CoRR}}  \bibinfo{volume}{abs/1809.02444}
  (\bibinfo{year}{2018}).
\newblock
\showeprint[arxiv]{1809.02444}
\urldef\tempurl%
\url{http://arxiv.org/abs/1809.02444}
\showURL{%
\tempurl}


\bibitem[\protect\citeauthoryear{Chen}{Chen}{2017}]%
        {scala-tensor}
\bibfield{author}{\bibinfo{person}{Tongfei Chen}.}
  \bibinfo{year}{2017}\natexlab{}.
\newblock \showarticletitle{Typesafe Abstractions for Tensor Operations}.
\newblock \bibinfo{journal}{\emph{CoRR}}  \bibinfo{volume}{abs/1710.06892}
  (\bibinfo{year}{2017}).
\newblock
\showeprint[arxiv]{1710.06892}
\urldef\tempurl%
\url{http://arxiv.org/abs/1710.06892}
\showURL{%
\tempurl}


\bibitem[\protect\citeauthoryear{Chen, Sahiner, Samuelson, Pezeshk, and
  Petrick}{Chen et~al\mbox{.}}{2018}]%
        {Chen2018Calibration}
\bibfield{author}{\bibinfo{person}{Weijie Chen}, \bibinfo{person}{Berkman
  Sahiner}, \bibinfo{person}{Frank~W. Samuelson}, \bibinfo{person}{Aria
  Pezeshk}, {and} \bibinfo{person}{Nicholas Petrick}.}
  \bibinfo{year}{2018}\natexlab{}.
\newblock \showarticletitle{Calibration of medical diagnostic classifier scores
  to the probability of disease}. In \bibinfo{booktitle}{\emph{Statistical
  methods in medical research}}.
\newblock


\bibitem[\protect\citeauthoryear{Chollet}{Chollet}{2015}]%
        {keras}
\bibfield{author}{\bibinfo{person}{François Chollet}.}
  \bibinfo{year}{2015}\natexlab{}.
\newblock \bibinfo{title}{Keras}.
\newblock \bibinfo{howpublished}{\url{https://github.com/fchollet/keras}}.
\newblock


\bibitem[\protect\citeauthoryear{Dolby, Shinnar, Allain, and Reinen}{Dolby
  et~al\mbox{.}}{2018}]%
        {Ariadne-ibm}
\bibfield{author}{\bibinfo{person}{Julian Dolby}, \bibinfo{person}{Avraham
  Shinnar}, \bibinfo{person}{Allison Allain}, {and} \bibinfo{person}{Jenna
  Reinen}.} \bibinfo{year}{2018}\natexlab{}.
\newblock \showarticletitle{Ariadne: Analysis for Machine Learning Programs}.
  In \bibinfo{booktitle}{\emph{Proceedings of the 2Nd ACM SIGPLAN International
  Workshop on Machine Learning and Programming Languages}}
  \emph{(\bibinfo{series}{MAPL 2018})}. \bibinfo{publisher}{ACM},
  \bibinfo{address}{New York, NY, USA}, \bibinfo{pages}{1--10}.
\newblock
\showISBNx{978-1-4503-5834-7}
\urldef\tempurl%
\url{https://doi.org/10.1145/3211346.3211349}
\showDOI{\tempurl}


\bibitem[\protect\citeauthoryear{Doshi-Velez and Kim}{Doshi-Velez and
  Kim}{2017}]%
        {DoshiVelez2017}
\bibfield{author}{\bibinfo{person}{Finale Doshi-Velez} {and}
  \bibinfo{person}{Been Kim}.} \bibinfo{year}{2017}\natexlab{}.
\newblock \showarticletitle{Towards A Rigorous Science of Interpretable Machine
  Learning}.
\newblock


\bibitem[\protect\citeauthoryear{Du, Xie, Li, Ma, Liu, and Zhao}{Du
  et~al\mbox{.}}{2019}]%
        {deepstellar}
\bibfield{author}{\bibinfo{person}{Xiaoning Du}, \bibinfo{person}{Xiaofei Xie},
  \bibinfo{person}{Yi Li}, \bibinfo{person}{Lei Ma}, \bibinfo{person}{Yang
  Liu}, {and} \bibinfo{person}{Jianjun Zhao}.} \bibinfo{year}{2019}\natexlab{}.
\newblock \showarticletitle{DeepStellar: Model-Based Quantitative Analysis of
  Stateful Deep Learning Systems}. In \bibinfo{booktitle}{\emph{Proceedings of
  the 2019 27th ACM Joint Meeting on European Software Engineering Conference
  and Symposium on the Foundations of Software Engineering}}
  \emph{(\bibinfo{series}{ESEC/FSE 2019})}. \bibinfo{publisher}{Association for
  Computing Machinery}, \bibinfo{address}{New York, NY, USA},
  \bibinfo{pages}{477–487}.
\newblock
\showISBNx{9781450355728}
\urldef\tempurl%
\url{https://doi.org/10.1145/3338906.3338954}
\showDOI{\tempurl}


\bibitem[\protect\citeauthoryear{Dwarakanath, Ahuja, Podder, Vinu, Naskar, and
  MV}{Dwarakanath et~al\mbox{.}}{2019}]%
        {meta-accenture2}
\bibfield{author}{\bibinfo{person}{Anurag Dwarakanath}, \bibinfo{person}{Manish
  Ahuja}, \bibinfo{person}{Sanjay Podder}, \bibinfo{person}{Silja Vinu},
  \bibinfo{person}{Arijit Naskar}, {and} \bibinfo{person}{Koushik MV}.}
  \bibinfo{year}{2019}\natexlab{}.
\newblock \showarticletitle{Metamorphic Testing of a Deep Learning Based
  Forecaster}. In \bibinfo{booktitle}{\emph{Proceedings of the 4th
  International Workshop on Metamorphic Testing}} \emph{(\bibinfo{series}{MET
  ’19})}. \bibinfo{publisher}{IEEE Press}, \bibinfo{pages}{40–47}.
\newblock
\urldef\tempurl%
\url{https://doi.org/10.1109/MET.2019.00014}
\showDOI{\tempurl}


\bibitem[\protect\citeauthoryear{Dwarakanath, Ahuja, Sikand, Rao, Bose, Dubash,
  and Podder}{Dwarakanath et~al\mbox{.}}{2018}]%
        {meta-accenture}
\bibfield{author}{\bibinfo{person}{Anurag Dwarakanath}, \bibinfo{person}{Manish
  Ahuja}, \bibinfo{person}{Samarth Sikand}, \bibinfo{person}{Raghotham~M. Rao},
  \bibinfo{person}{R.~P. Jagadeesh~Chandra Bose}, \bibinfo{person}{Neville
  Dubash}, {and} \bibinfo{person}{Sanjay Podder}.}
  \bibinfo{year}{2018}\natexlab{}.
\newblock \showarticletitle{Identifying Implementation Bugs in Machine Learning
  based Image Classifiers using Metamorphic Testing}.
\newblock \bibinfo{journal}{\emph{CoRR}}  \bibinfo{volume}{abs/1808.05353}
  (\bibinfo{year}{2018}).
\newblock


\bibitem[\protect\citeauthoryear{Dwork, Hardt, Pitassi, Reingold, and
  Zemel}{Dwork et~al\mbox{.}}{2011}]%
        {individual-fair}
\bibfield{author}{\bibinfo{person}{Cynthia Dwork}, \bibinfo{person}{Moritz
  Hardt}, \bibinfo{person}{Toniann Pitassi}, \bibinfo{person}{Omer Reingold},
  {and} \bibinfo{person}{Richard~S. Zemel}.} \bibinfo{year}{2011}\natexlab{}.
\newblock \showarticletitle{Fairness Through Awareness}.
\newblock \bibinfo{journal}{\emph{CoRR}} (\bibinfo{year}{2011}).
\newblock
\urldef\tempurl%
\url{http://arxiv.org/abs/1104.3913}
\showURL{%
\tempurl}


\bibitem[\protect\citeauthoryear{Ferreira, Silva, and Valente}{Ferreira
  et~al\mbox{.}}{2019}]%
        {ferreira2019software}
\bibfield{author}{\bibinfo{person}{Fabio Ferreira},
  \bibinfo{person}{Luciana~Lourdes Silva}, {and} \bibinfo{person}{Marco~Tulio
  Valente}.} \bibinfo{year}{2019}\natexlab{}.
\newblock \bibinfo{title}{Software Engineering Meets Deep Learning: A
  Literature Review}.
\newblock
\newblock
\showeprint[arxiv]{cs.SE/1909.11436}


\bibitem[\protect\citeauthoryear{Galhotra, Brun, and Meliou}{Galhotra
  et~al\mbox{.}}{2017}]%
        {themis}
\bibfield{author}{\bibinfo{person}{Sainyam Galhotra}, \bibinfo{person}{Yuriy
  Brun}, {and} \bibinfo{person}{Alexandra Meliou}.}
  \bibinfo{year}{2017}\natexlab{}.
\newblock \showarticletitle{Fairness Testing: Testing Software for
  Discrimination}. In \bibinfo{booktitle}{\emph{Proceedings of the 2017 11th
  Joint Meeting on Foundations of Software Engineering}}
  \emph{(\bibinfo{series}{ESEC/FSE 2017})}. \bibinfo{publisher}{ACM},
  \bibinfo{address}{New York, NY, USA}, \bibinfo{pages}{498--510}.
\newblock
\showISBNx{978-1-4503-5105-8}
\urldef\tempurl%
\url{https://doi.org/10.1145/3106237.3106277}
\showDOI{\tempurl}


\bibitem[\protect\citeauthoryear{Hall, Frank, Holmes, Pfahringer, Reutemann,
  and Witten}{Hall et~al\mbox{.}}{2009}]%
        {weka}
\bibfield{author}{\bibinfo{person}{Mark Hall}, \bibinfo{person}{Eibe Frank},
  \bibinfo{person}{Geoffrey Holmes}, \bibinfo{person}{Bernhard Pfahringer},
  \bibinfo{person}{Peter Reutemann}, {and} \bibinfo{person}{Ian~H. Witten}.}
  \bibinfo{year}{2009}\natexlab{}.
\newblock \showarticletitle{The WEKA Data Mining Software: An Update}.
\newblock \bibinfo{journal}{\emph{SIGKDD Explor. Newsl.}} \bibinfo{volume}{11},
  \bibinfo{number}{1} (\bibinfo{date}{Nov.} \bibinfo{year}{2009}),
  \bibinfo{pages}{10--18}.
\newblock
\showISSN{1931-0145}
\urldef\tempurl%
\url{https://doi.org/10.1145/1656274.1656278}
\showDOI{\tempurl}


\bibitem[\protect\citeauthoryear{Hohman, Kahng, Pienta, and Chau}{Hohman
  et~al\mbox{.}}{2018}]%
        {visualize3}
\bibfield{author}{\bibinfo{person}{Fred Hohman}, \bibinfo{person}{Minsuk
  Kahng}, \bibinfo{person}{Robert Pienta}, {and} \bibinfo{person}{Duen~Horng
  Chau}.} \bibinfo{year}{2018}\natexlab{}.
\newblock \showarticletitle{Visual Analytics in Deep Learning: An Interrogative
  Survey for the Next Frontiers}.
\newblock \bibinfo{journal}{\emph{IEEE Transactions on Visualization and
  Computer Graphics}}  \bibinfo{volume}{25} (\bibinfo{year}{2018}),
  \bibinfo{pages}{2674--2693}.
\newblock


\bibitem[\protect\citeauthoryear{Islam, Nguyen, Pan, and Rajan}{Islam
  et~al\mbox{.}}{2019}]%
        {fse-bug-collection}
\bibfield{author}{\bibinfo{person}{Md~Johirul Islam}, \bibinfo{person}{Giang
  Nguyen}, \bibinfo{person}{Rangeet Pan}, {and} \bibinfo{person}{Hridesh
  Rajan}.} \bibinfo{year}{2019}\natexlab{}.
\newblock \showarticletitle{A Comprehensive Study on Deep Learning Bug
  Characteristics}. In \bibinfo{booktitle}{\emph{Proceedings of the 2019 27th
  ACM Joint Meeting on European Software Engineering Conference and Symposium
  on the Foundations of Software Engineering}} \emph{(\bibinfo{series}{ESEC/FSE
  2019})}. \bibinfo{publisher}{ACM}, \bibinfo{address}{New York, NY, USA},
  \bibinfo{pages}{510--520}.
\newblock
\showISBNx{978-1-4503-5572-8}
\urldef\tempurl%
\url{https://doi.org/10.1145/3338906.3338955}
\showDOI{\tempurl}


\bibitem[\protect\citeauthoryear{Jia, Shelhamer, Donahue, Karayev, Long,
  Girshick, Guadarrama, and Darrell}{Jia et~al\mbox{.}}{2014}]%
        {caffe-paper}
\bibfield{author}{\bibinfo{person}{Yangqing Jia}, \bibinfo{person}{Evan
  Shelhamer}, \bibinfo{person}{Jeff Donahue}, \bibinfo{person}{Sergey Karayev},
  \bibinfo{person}{Jonathan Long}, \bibinfo{person}{Ross Girshick},
  \bibinfo{person}{Sergio Guadarrama}, {and} \bibinfo{person}{Trevor Darrell}.}
  \bibinfo{year}{2014}\natexlab{}.
\newblock \showarticletitle{Caffe: Convolutional Architecture for Fast Feature
  Embedding}. In \bibinfo{booktitle}{\emph{Proceedings of the 22Nd ACM
  International Conference on Multimedia}} \emph{(\bibinfo{series}{MM '14})}.
  \bibinfo{publisher}{ACM}, \bibinfo{address}{New York, NY, USA},
  \bibinfo{pages}{675--678}.
\newblock
\showISBNx{978-1-4503-3063-3}
\urldef\tempurl%
\url{https://doi.org/10.1145/2647868.2654889}
\showDOI{\tempurl}


\bibitem[\protect\citeauthoryear{Kim, Feldt, and Yoo}{Kim
  et~al\mbox{.}}{2018}]%
        {surprise}
\bibfield{author}{\bibinfo{person}{Jinhan Kim}, \bibinfo{person}{Robert Feldt},
  {and} \bibinfo{person}{Shin Yoo}.} \bibinfo{year}{2018}\natexlab{}.
\newblock \showarticletitle{Guiding Deep Learning System Testing using Surprise
  Adequacy}.
\newblock \bibinfo{journal}{\emph{CoRR}}  \bibinfo{volume}{abs/1808.08444}
  (\bibinfo{year}{2018}).
\newblock
\showeprint[arxiv]{1808.08444}
\urldef\tempurl%
\url{http://arxiv.org/abs/1808.08444}
\showURL{%
\tempurl}


\bibitem[\protect\citeauthoryear{Krishnan, Franklin, Goldberg, and Wu}{Krishnan
  et~al\mbox{.}}{2017}]%
        {boostclean}
\bibfield{author}{\bibinfo{person}{Sanjay Krishnan},
  \bibinfo{person}{Michael~J. Franklin}, \bibinfo{person}{Ken Goldberg}, {and}
  \bibinfo{person}{Eugene Wu}.} \bibinfo{year}{2017}\natexlab{}.
\newblock \showarticletitle{BoostClean: Automated Error Detection and Repair
  for Machine Learning}.
\newblock \bibinfo{journal}{\emph{CoRR}}  \bibinfo{volume}{abs/1711.01299}
  (\bibinfo{year}{2017}).
\newblock
\showeprint[arxiv]{1711.01299}
\urldef\tempurl%
\url{http://arxiv.org/abs/1711.01299}
\showURL{%
\tempurl}


\bibitem[\protect\citeauthoryear{Lagouvardos, Dolby, Grech, Antoniadis, and
  Smaragdakis}{Lagouvardos et~al\mbox{.}}{2020}]%
        {Pythia-ibm}
\bibfield{author}{\bibinfo{person}{Sifis Lagouvardos}, \bibinfo{person}{Julian
  Dolby}, \bibinfo{person}{Neville Grech}, \bibinfo{person}{Anastasios
  Antoniadis}, {and} \bibinfo{person}{Yannis Smaragdakis}.}
  \bibinfo{year}{2020}\natexlab{}.
\newblock \showarticletitle{Static {Analysis} of {Shape} in {TensorFlow}
  {Programs}}. In \bibinfo{booktitle}{\emph{34th European Conference on
  Object-Oriented Programming (ECOOP 2020)}} \emph{(\bibinfo{series}{Leibniz
  International Proceedings in Informatics (LIPIcs)})}.
  \bibinfo{publisher}{Schloss Dagstuhl--Leibniz-Zentrum fuer Informatik},
  \bibinfo{address}{Dagstuhl, Germany}, 30.
\newblock
\urldef\tempurl%
\url{https://doi.org/10.4230/LIPIcs.ECOOP.2020.15}
\showDOI{\tempurl}


\bibitem[\protect\citeauthoryear{LeCun and Cortes}{LeCun and Cortes}{2010}]%
        {lecun-mnisthandwrittendigit-2010}
\bibfield{author}{\bibinfo{person}{Yann LeCun} {and} \bibinfo{person}{Corinna
  Cortes}.} \bibinfo{year}{2010}\natexlab{}.
\newblock \showarticletitle{{MNIST} handwritten digit database}.
\newblock \bibinfo{howpublished}{http://yann.lecun.com/exdb/mnist/}.
\newblock  (\bibinfo{year}{2010}).
\newblock
\urldef\tempurl%
\url{http://yann.lecun.com/exdb/mnist/}
\showURL{%
\tempurl}


\bibitem[\protect\citeauthoryear{Liu, Luo, Wang, and Tang}{Liu
  et~al\mbox{.}}{2015}]%
        {celebA_dataset}
\bibfield{author}{\bibinfo{person}{Ziwei Liu}, \bibinfo{person}{Ping Luo},
  \bibinfo{person}{Xiaogang Wang}, {and} \bibinfo{person}{Xiaoou Tang}.}
  \bibinfo{year}{2015}\natexlab{}.
\newblock \showarticletitle{Deep Learning Face Attributes in the Wild}. In
  \bibinfo{booktitle}{\emph{Proceedings of International Conference on Computer
  Vision (ICCV)}}.
\newblock


\bibitem[\protect\citeauthoryear{{Ma}, {Juefei-Xu}, {Xue}, {Li}, {Li}, {Liu},
  and {Zhao}}{{Ma} et~al\mbox{.}}{2019}]%
        {deepct}
\bibfield{author}{\bibinfo{person}{L. {Ma}}, \bibinfo{person}{F. {Juefei-Xu}},
  \bibinfo{person}{M. {Xue}}, \bibinfo{person}{B. {Li}}, \bibinfo{person}{L.
  {Li}}, \bibinfo{person}{Y. {Liu}}, {and} \bibinfo{person}{J. {Zhao}}.}
  \bibinfo{year}{2019}\natexlab{}.
\newblock \showarticletitle{DeepCT: Tomographic Combinatorial Testing for Deep
  Learning Systems}. In \bibinfo{booktitle}{\emph{2019 IEEE 26th International
  Conference on Software Analysis, Evolution and Reengineering (SANER)}}.
  \bibinfo{pages}{614--618}.
\newblock
\urldef\tempurl%
\url{https://doi.org/10.1109/SANER.2019.8668044}
\showDOI{\tempurl}


\bibitem[\protect\citeauthoryear{Ma, Juefei-Xu, Zhang, Sun, Xue, Li, Chen, Su,
  Li, Liu, Zhao, and Wang}{Ma et~al\mbox{.}}{2018a}]%
        {deepguage}
\bibfield{author}{\bibinfo{person}{Lei Ma}, \bibinfo{person}{Felix Juefei-Xu},
  \bibinfo{person}{Fuyuan Zhang}, \bibinfo{person}{Jiyuan Sun},
  \bibinfo{person}{Minhui Xue}, \bibinfo{person}{Bo Li},
  \bibinfo{person}{Chunyang Chen}, \bibinfo{person}{Ting Su},
  \bibinfo{person}{Li Li}, \bibinfo{person}{Yang Liu}, \bibinfo{person}{Jianjun
  Zhao}, {and} \bibinfo{person}{Yadong Wang}.}
  \bibinfo{year}{2018}\natexlab{a}.
\newblock \showarticletitle{DeepGauge: Multi-granularity Testing Criteria for
  Deep Learning Systems}. In \bibinfo{booktitle}{\emph{Proceedings of the 33rd
  ACM/IEEE International Conference on Automated Software Engineering}}
  \emph{(\bibinfo{series}{ASE 2018})}. \bibinfo{publisher}{ACM},
  \bibinfo{address}{New York, NY, USA}, \bibinfo{pages}{120--131}.
\newblock
\showISBNx{978-1-4503-5937-5}
\urldef\tempurl%
\url{https://doi.org/10.1145/3238147.3238202}
\showDOI{\tempurl}


\bibitem[\protect\citeauthoryear{Ma, Zhang, Sun, Xue, Li, Juefei-Xu, Xie, Li,
  Liu, Zhao, and Wang}{Ma et~al\mbox{.}}{2018c}]%
        {DeepMutationMT}
\bibfield{author}{\bibinfo{person}{Lei Ma}, \bibinfo{person}{Fuyuan Zhang},
  \bibinfo{person}{Jiyuan Sun}, \bibinfo{person}{Minhui Xue},
  \bibinfo{person}{Bao~Qin Li}, \bibinfo{person}{Felix Juefei-Xu},
  \bibinfo{person}{Chao Xie}, \bibinfo{person}{Li Li}, \bibinfo{person}{Yang~P.
  Liu}, \bibinfo{person}{Jianjun Zhao}, {and} \bibinfo{person}{Yadong Wang}.}
  \bibinfo{year}{2018}\natexlab{c}.
\newblock \showarticletitle{DeepMutation: Mutation Testing of Deep Learning
  Systems}.
\newblock \bibinfo{journal}{\emph{2018 IEEE 29th International Symposium on
  Software Reliability Engineering (ISSRE)}} (\bibinfo{year}{2018}),
  \bibinfo{pages}{100--111}.
\newblock


\bibitem[\protect\citeauthoryear{Ma, Liu, Lee, Zhang, and Grama}{Ma
  et~al\mbox{.}}{2018b}]%
        {mode}
\bibfield{author}{\bibinfo{person}{Shiqing Ma}, \bibinfo{person}{Yingqi Liu},
  \bibinfo{person}{Wen-Chuan Lee}, \bibinfo{person}{Xiangyu Zhang}, {and}
  \bibinfo{person}{Ananth Grama}.} \bibinfo{year}{2018}\natexlab{b}.
\newblock \showarticletitle{MODE: Automated Neural Network Model Debugging via
  State Differential Analysis and Input Selection}. In
  \bibinfo{booktitle}{\emph{Proceedings of the 2018 26th ACM Joint Meeting on
  European Software Engineering Conference and Symposium on the Foundations of
  Software Engineering}} \emph{(\bibinfo{series}{ESEC/FSE 2018})}.
  \bibinfo{publisher}{ACM}, \bibinfo{address}{New York, NY, USA},
  \bibinfo{pages}{175--186}.
\newblock
\showISBNx{978-1-4503-5573-5}
\urldef\tempurl%
\url{https://doi.org/10.1145/3236024.3236082}
\showDOI{\tempurl}


\bibitem[\protect\citeauthoryear{Moosavi{-}Dezfooli, Fawzi, and
  Frossard}{Moosavi{-}Dezfooli et~al\mbox{.}}{2015}]%
        {deepfool}
\bibfield{author}{\bibinfo{person}{Seyed{-}Mohsen Moosavi{-}Dezfooli},
  \bibinfo{person}{Alhussein Fawzi}, {and} \bibinfo{person}{Pascal Frossard}.}
  \bibinfo{year}{2015}\natexlab{}.
\newblock \showarticletitle{DeepFool: a simple and accurate method to fool deep
  neural networks}.
\newblock \bibinfo{journal}{\emph{CoRR}}  \bibinfo{volume}{abs/1511.04599}
  (\bibinfo{year}{2015}).
\newblock
\showeprint[arxiv]{1511.04599}
\urldef\tempurl%
\url{http://arxiv.org/abs/1511.04599}
\showURL{%
\tempurl}


\bibitem[\protect\citeauthoryear{Murphy, Kaiser, and Arias}{Murphy
  et~al\mbox{.}}{2007}]%
        {Murphy_anapproach}
\bibfield{author}{\bibinfo{person}{Chris Murphy}, \bibinfo{person}{Gail
  Kaiser}, {and} \bibinfo{person}{Marta Arias}.}
  \bibinfo{year}{2007}\natexlab{}.
\newblock \showarticletitle{An Approach to Software Testing of Machine Learning
  Applications}.
\newblock


\bibitem[\protect\citeauthoryear{Odena, Olsson, Andersen, and Goodfellow}{Odena
  et~al\mbox{.}}{2019}]%
        {tensorfuzz}
\bibfield{author}{\bibinfo{person}{Augustus Odena}, \bibinfo{person}{Catherine
  Olsson}, \bibinfo{person}{David Andersen}, {and} \bibinfo{person}{Ian
  Goodfellow}.} \bibinfo{year}{2019}\natexlab{}.
\newblock \showarticletitle{{T}ensor{F}uzz: Debugging Neural Networks with
  Coverage-Guided Fuzzing}. In \bibinfo{booktitle}{\emph{Proceedings of the
  36th International Conference on Machine Learning}}
  \emph{(\bibinfo{series}{Proceedings of Machine Learning Research})},
  \bibfield{editor}{\bibinfo{person}{Kamalika Chaudhuri} {and}
  \bibinfo{person}{Ruslan Salakhutdinov}} (Eds.), Vol.~\bibinfo{volume}{97}.
  \bibinfo{publisher}{PMLR}, \bibinfo{address}{Long Beach, California, USA},
  \bibinfo{pages}{4901--4911}.
\newblock
\urldef\tempurl%
\url{http://proceedings.mlr.press/v97/odena19a.html}
\showURL{%
\tempurl}


\bibitem[\protect\citeauthoryear{Paszke, Gross, Chintala, Chanan, Yang, DeVito,
  Lin, Desmaison, Antiga, and Lerer}{Paszke et~al\mbox{.}}{2017}]%
        {pyTorch-paper}
\bibfield{author}{\bibinfo{person}{Adam Paszke}, \bibinfo{person}{Sam Gross},
  \bibinfo{person}{Soumith Chintala}, \bibinfo{person}{Gregory Chanan},
  \bibinfo{person}{Edward Yang}, \bibinfo{person}{Zachary DeVito},
  \bibinfo{person}{Zeming Lin}, \bibinfo{person}{Alban Desmaison},
  \bibinfo{person}{Luca Antiga}, {and} \bibinfo{person}{Adam Lerer}.}
  \bibinfo{year}{2017}\natexlab{}.
\newblock \showarticletitle{Automatic differentiation in PyTorch}. In
  \bibinfo{booktitle}{\emph{NIPS-W}}.
\newblock


\bibitem[\protect\citeauthoryear{Pei, Cao, Yang, and Jana}{Pei
  et~al\mbox{.}}{2017}]%
        {deepxplore}
\bibfield{author}{\bibinfo{person}{Kexin Pei}, \bibinfo{person}{Yinzhi Cao},
  \bibinfo{person}{Junfeng Yang}, {and} \bibinfo{person}{Suman Jana}.}
  \bibinfo{year}{2017}\natexlab{}.
\newblock \showarticletitle{DeepXplore: Automated Whitebox Testing of Deep
  Learning Systems}. In \bibinfo{booktitle}{\emph{Proceedings of the 26th
  Symposium on Operating Systems Principles}} \emph{(\bibinfo{series}{SOSP
  '17})}. \bibinfo{publisher}{ACM}, \bibinfo{address}{New York, NY, USA},
  \bibinfo{pages}{1--18}.
\newblock
\showISBNx{978-1-4503-5085-3}
\urldef\tempurl%
\url{https://doi.org/10.1145/3132747.3132785}
\showDOI{\tempurl}


\bibitem[\protect\citeauthoryear{Pham, Lutellier, Qi, and Tan}{Pham
  et~al\mbox{.}}{2019}]%
        {cradle}
\bibfield{author}{\bibinfo{person}{Hung~Viet Pham}, \bibinfo{person}{Thibaud
  Lutellier}, \bibinfo{person}{Weizhen Qi}, {and} \bibinfo{person}{Lin Tan}.}
  \bibinfo{year}{2019}\natexlab{}.
\newblock \showarticletitle{CRADLE: Cross-backend Validation to Detect and
  Localize Bugs in Deep Learning Libraries}. In
  \bibinfo{booktitle}{\emph{Proceedings of the 41st International Conference on
  Software Engineering}} \emph{(\bibinfo{series}{ICSE '19})}.
  \bibinfo{publisher}{IEEE Press}, \bibinfo{address}{Piscataway, NJ, USA},
  \bibinfo{pages}{1027--1038}.
\newblock
\urldef\tempurl%
\url{https://doi.org/10.1109/ICSE.2019.00107}
\showDOI{\tempurl}


\bibitem[\protect\citeauthoryear{Rossum}{Rossum}{1995}]%
        {python_reference}
\bibfield{author}{\bibinfo{person}{Guido Rossum}.}
  \bibinfo{year}{1995}\natexlab{}.
\newblock \bibinfo{booktitle}{\emph{Python Reference Manual}}.
\newblock \bibinfo{type}{{T}echnical {R}eport}. \bibinfo{address}{Amsterdam,
  The Netherlands, The Netherlands}.
\newblock


\bibitem[\protect\citeauthoryear{Sridharan, Chandra, Dolby, Fink, and
  Yahav}{Sridharan et~al\mbox{.}}{2013}]%
        {wala-page}
\bibfield{author}{\bibinfo{person}{Manu Sridharan}, \bibinfo{person}{Satish
  Chandra}, \bibinfo{person}{Julian Dolby}, \bibinfo{person}{Stephen~J. Fink},
  {and} \bibinfo{person}{Eran Yahav}.} \bibinfo{year}{2013}\natexlab{}.
\newblock \bibinfo{booktitle}{\emph{Alias Analysis for Object-Oriented
  Programs}}.
\newblock \bibinfo{publisher}{Springer-Verlag}, \bibinfo{address}{Berlin,
  Heidelberg}, \bibinfo{pages}{196–232}.
\newblock
\showISBNx{9783642369452}


\bibitem[\protect\citeauthoryear{Tansley}{Tansley}{2000}]%
        {unix_utilities}
\bibfield{author}{\bibinfo{person}{David Tansley}.}
  \bibinfo{year}{2000}\natexlab{}.
\newblock \bibinfo{booktitle}{\emph{Linux and Unix Shell Programming}}.
\newblock \bibinfo{publisher}{Addison-Wesley Longman Publishing Co., Inc.},
  \bibinfo{address}{Boston, MA, USA}.
\newblock
\showISBNx{0-201-67472-6}


\bibitem[\protect\citeauthoryear{Tian, Pei, Jana, and Ray}{Tian
  et~al\mbox{.}}{2018}]%
        {deeptest}
\bibfield{author}{\bibinfo{person}{Yuchi Tian}, \bibinfo{person}{Kexin Pei},
  \bibinfo{person}{Suman Jana}, {and} \bibinfo{person}{Baishakhi Ray}.}
  \bibinfo{year}{2018}\natexlab{}.
\newblock \showarticletitle{DeepTest: Automated Testing of
  Deep-neural-network-driven Autonomous Cars}. In
  \bibinfo{booktitle}{\emph{Proceedings of the 40th International Conference on
  Software Engineering}} \emph{(\bibinfo{series}{ICSE '18})}.
  \bibinfo{publisher}{ACM}, \bibinfo{address}{New York, NY, USA},
  \bibinfo{pages}{303--314}.
\newblock
\showISBNx{978-1-4503-5638-1}
\urldef\tempurl%
\url{https://doi.org/10.1145/3180155.3180220}
\showDOI{\tempurl}


\bibitem[\protect\citeauthoryear{Udeshi, Arora, and Chattopadhyay}{Udeshi
  et~al\mbox{.}}{2018}]%
        {aequitas}
\bibfield{author}{\bibinfo{person}{Sakshi Udeshi}, \bibinfo{person}{Pryanshu
  Arora}, {and} \bibinfo{person}{Sudipta Chattopadhyay}.}
  \bibinfo{year}{2018}\natexlab{}.
\newblock \showarticletitle{Automated Directed Fairness Testing}. In
  \bibinfo{booktitle}{\emph{Proceedings of the 33rd ACM/IEEE International
  Conference on Automated Software Engineering}} \emph{(\bibinfo{series}{ASE
  2018})}. \bibinfo{publisher}{ACM}, \bibinfo{address}{New York, NY, USA},
  \bibinfo{pages}{98--108}.
\newblock
\showISBNx{978-1-4503-5937-5}
\urldef\tempurl%
\url{https://doi.org/10.1145/3238147.3238165}
\showDOI{\tempurl}


\bibitem[\protect\citeauthoryear{Vartak, F.~da Trindade, Madden, and
  Zaharia}{Vartak et~al\mbox{.}}{2018}]%
        {Vartak:2018}
\bibfield{author}{\bibinfo{person}{Manasi Vartak}, \bibinfo{person}{Joana~M.
  F.~da Trindade}, \bibinfo{person}{Samuel Madden}, {and}
  \bibinfo{person}{Matei Zaharia}.} \bibinfo{year}{2018}\natexlab{}.
\newblock \showarticletitle{MISTIQUE: A System to Store and Query Model
  Intermediates for Model Diagnosis}. In \bibinfo{booktitle}{\emph{Proceedings
  of the 2018 International Conference on Management of Data}}
  \emph{(\bibinfo{series}{SIGMOD '18})}. \bibinfo{publisher}{ACM},
  \bibinfo{address}{New York, NY, USA}, \bibinfo{pages}{1285--1300}.
\newblock
\showISBNx{978-1-4503-4703-7}
\urldef\tempurl%
\url{https://doi.org/10.1145/3183713.3196934}
\showDOI{\tempurl}


\bibitem[\protect\citeauthoryear{{Verma} and {Rubin}}{{Verma} and
  {Rubin}}{2018}]%
        {fairness-definitions}
\bibfield{author}{\bibinfo{person}{S. {Verma}} {and} \bibinfo{person}{J.
  {Rubin}}.} \bibinfo{year}{2018}\natexlab{}.
\newblock \showarticletitle{Fairness Definitions Explained}. In
  \bibinfo{booktitle}{\emph{2018 IEEE/ACM International Workshop on Software
  Fairness (FairWare)}}. \bibinfo{pages}{1--7}.
\newblock
\urldef\tempurl%
\url{https://doi.org/10.23919/FAIRWARE.2018.8452913}
\showDOI{\tempurl}


\bibitem[\protect\citeauthoryear{{Wongsuphasawat}, {Smilkov}, {Wexler},
  {Wilson}, {Mané}, {Fritz}, {Krishnan}, {Viégas}, and
  {Wattenberg}}{{Wongsuphasawat} et~al\mbox{.}}{2018}]%
        {visualize1}
\bibfield{author}{\bibinfo{person}{K. {Wongsuphasawat}}, \bibinfo{person}{D.
  {Smilkov}}, \bibinfo{person}{J. {Wexler}}, \bibinfo{person}{J. {Wilson}},
  \bibinfo{person}{D. {Mané}}, \bibinfo{person}{D. {Fritz}},
  \bibinfo{person}{D. {Krishnan}}, \bibinfo{person}{F.~B. {Viégas}}, {and}
  \bibinfo{person}{M. {Wattenberg}}.} \bibinfo{year}{2018}\natexlab{}.
\newblock \showarticletitle{Visualizing Dataflow Graphs of Deep Learning Models
  in TensorFlow}.
\newblock \bibinfo{journal}{\emph{IEEE Transactions on Visualization and
  Computer Graphics}} \bibinfo{volume}{24}, \bibinfo{number}{1}
  (\bibinfo{date}{Jan} \bibinfo{year}{2018}), \bibinfo{pages}{1--12}.
\newblock
\urldef\tempurl%
\url{https://doi.org/10.1109/TVCG.2017.2744878}
\showDOI{\tempurl}


\bibitem[\protect\citeauthoryear{{Xiao}, {Li}, {Zhang}, and {Xu}}{{Xiao}
  et~al\mbox{.}}{2018}]%
        {security-problem}
\bibfield{author}{\bibinfo{person}{Q. {Xiao}}, \bibinfo{person}{K. {Li}},
  \bibinfo{person}{D. {Zhang}}, {and} \bibinfo{person}{W. {Xu}}.}
  \bibinfo{year}{2018}\natexlab{}.
\newblock \showarticletitle{Security Risks in Deep Learning Implementations}.
  In \bibinfo{booktitle}{\emph{2018 IEEE Security and Privacy Workshops
  (SPW)}}. \bibinfo{pages}{123--128}.
\newblock
\urldef\tempurl%
\url{https://doi.org/10.1109/SPW.2018.00027}
\showDOI{\tempurl}


\bibitem[\protect\citeauthoryear{Xie, Ho, Murphy, Kaiser, Xu, and Chen}{Xie
  et~al\mbox{.}}{2011}]%
        {Xie:2011}
\bibfield{author}{\bibinfo{person}{Xiaoyuan Xie}, \bibinfo{person}{Joshua W.~K.
  Ho}, \bibinfo{person}{Christian Murphy}, \bibinfo{person}{Gail Kaiser},
  \bibinfo{person}{Baowen Xu}, {and} \bibinfo{person}{Tsong~Yueh Chen}.}
  \bibinfo{year}{2011}\natexlab{}.
\newblock \showarticletitle{Testing and Validating Machine Learning Classifiers
  by Metamorphic Testing}.
\newblock \bibinfo{journal}{\emph{J. Syst. Softw.}} \bibinfo{volume}{84},
  \bibinfo{number}{4} (\bibinfo{date}{April} \bibinfo{year}{2011}),
  \bibinfo{pages}{544--558}.
\newblock
\showISSN{0164-1212}
\urldef\tempurl%
\url{https://doi.org/10.1016/j.jss.2010.11.920}
\showDOI{\tempurl}


\bibitem[\protect\citeauthoryear{Xie, Ma, Juefei-Xu, Xue, Chen, Liu, Zhao, Li,
  Yin, and See}{Xie et~al\mbox{.}}{2019}]%
        {deephunter}
\bibfield{author}{\bibinfo{person}{Xiaofei Xie}, \bibinfo{person}{Lei Ma},
  \bibinfo{person}{Felix Juefei-Xu}, \bibinfo{person}{Minhui Xue},
  \bibinfo{person}{Hongxu Chen}, \bibinfo{person}{Yang Liu},
  \bibinfo{person}{Jianjun Zhao}, \bibinfo{person}{Bo Li},
  \bibinfo{person}{Jianxiong Yin}, {and} \bibinfo{person}{Simon See}.}
  \bibinfo{year}{2019}\natexlab{}.
\newblock \showarticletitle{DeepHunter: A Coverage-Guided Fuzz Testing
  Framework for Deep Neural Networks}. In \bibinfo{booktitle}{\emph{Proceedings
  of the 28th ACM SIGSOFT International Symposium on Software Testing and
  Analysis}} \emph{(\bibinfo{series}{ISSTA 2019})}.
  \bibinfo{publisher}{Association for Computing Machinery},
  \bibinfo{address}{New York, NY, USA}, \bibinfo{pages}{146–157}.
\newblock
\showISBNx{9781450362245}
\urldef\tempurl%
\url{https://doi.org/10.1145/3293882.3330579}
\showDOI{\tempurl}


\bibitem[\protect\citeauthoryear{Zhang, Harman, Ma, and Liu}{Zhang
  et~al\mbox{.}}{2019}]%
        {mlt-survey}
\bibfield{author}{\bibinfo{person}{Jie~M. Zhang}, \bibinfo{person}{Mark
  Harman}, \bibinfo{person}{Lei Ma}, {and} \bibinfo{person}{Yang Liu}.}
  \bibinfo{year}{2019}\natexlab{}.
\newblock \showarticletitle{Machine Learning Testing: Survey, Landscapes and
  Horizons}.
\newblock \bibinfo{journal}{\emph{CoRR}}  \bibinfo{volume}{abs/1906.10742}
  (\bibinfo{year}{2019}).
\newblock
\showeprint[arxiv]{1906.10742}
\urldef\tempurl%
\url{http://arxiv.org/abs/1906.10742}
\showURL{%
\tempurl}


\bibitem[\protect\citeauthoryear{Zhang, Zhang, Zhang, Liu, and Khurshid}{Zhang
  et~al\mbox{.}}{2018b}]%
        {deeproad}
\bibfield{author}{\bibinfo{person}{Mengshi Zhang}, \bibinfo{person}{Yuqun
  Zhang}, \bibinfo{person}{Lingming Zhang}, \bibinfo{person}{Cong Liu}, {and}
  \bibinfo{person}{Sarfraz Khurshid}.} \bibinfo{year}{2018}\natexlab{b}.
\newblock \showarticletitle{DeepRoad: GAN-based Metamorphic Testing and Input
  Validation Framework for Autonomous Driving Systems}. In
  \bibinfo{booktitle}{\emph{Proceedings of the 33rd ACM/IEEE International
  Conference on Automated Software Engineering}} \emph{(\bibinfo{series}{ASE
  2018})}. \bibinfo{publisher}{ACM}, \bibinfo{address}{New York, NY, USA},
  \bibinfo{pages}{132--142}.
\newblock
\showISBNx{978-1-4503-5937-5}
\urldef\tempurl%
\url{https://doi.org/10.1145/3238147.3238187}
\showDOI{\tempurl}


\bibitem[\protect\citeauthoryear{Zhang, Chen, Cheung, Xiong, and Zhang}{Zhang
  et~al\mbox{.}}{2018a}]%
        {issta-bug-collection}
\bibfield{author}{\bibinfo{person}{Yuhao Zhang}, \bibinfo{person}{Yifan Chen},
  \bibinfo{person}{Shing-Chi Cheung}, \bibinfo{person}{Yingfei Xiong}, {and}
  \bibinfo{person}{Lu Zhang}.} \bibinfo{year}{2018}\natexlab{a}.
\newblock \showarticletitle{An Empirical Study on TensorFlow Program Bugs}. In
  \bibinfo{booktitle}{\emph{Proceedings of the 27th ACM SIGSOFT International
  Symposium on Software Testing and Analysis}} \emph{(\bibinfo{series}{ISSTA
  2018})}. \bibinfo{publisher}{ACM}, \bibinfo{address}{New York, NY, USA},
  \bibinfo{pages}{129--140}.
\newblock
\showISBNx{978-1-4503-5699-2}
\urldef\tempurl%
\url{https://doi.org/10.1145/3213846.3213866}
\showDOI{\tempurl}


\bibitem[\protect\citeauthoryear{Zhou and Sun}{Zhou and Sun}{2018}]%
        {Zhou2018MetamorphicTF}
\bibfield{author}{\bibinfo{person}{Zhi~Quan Zhou} {and} \bibinfo{person}{Liqun
  Sun}.} \bibinfo{year}{2018}\natexlab{}.
\newblock \showarticletitle{Metamorphic Testing for Machine Translations:
  MT4MT}.
\newblock \bibinfo{journal}{\emph{2018 25th Australasian Software Engineering
  Conference (ASWEC)}} (\bibinfo{year}{2018}), \bibinfo{pages}{96--100}.
\newblock


\end{thebibliography}

\end{document}